\documentclass{article}

\usepackage[preprint]{neurips_2026}

\usepackage[utf8]{inputenc} 
\DeclareUnicodeCharacter{2212}{\ensuremath{-}}
\usepackage[T1]{fontenc}    
\usepackage{hyperref}       
\usepackage{url}            
\usepackage{booktabs}       
\usepackage{amsfonts}       
\usepackage{nicefrac}       
\usepackage{microtype}      
\usepackage{graphicx}
\usepackage{xcolor}         
\usepackage{placeins}
\usepackage{amsmath}
\usepackage{float}
\usepackage{subcaption}

\title{Same Model, Different Weakness: How Language and Modality Reshape the Jailbreak Attack Surface in Frontier MLLMs}

\author{%
  Casey Ford\thanks{Corresponding author} \And
Madison Van Doren \And
Sicheng Jin \And
Emily Dix \\
Appen \\
\texttt{\{cford, mvandoren, sjin, edix\}@appen.com} \\
  }
  
\begin{document}

\maketitle

\begin{abstract}
The attack surface of a multimodal large language model (MLLM) is language-dependent in ways that reveal the mechanistic structure of alignment failures. We present the first systematic cross-lingual, multimodal red-teaming study comparing jailbreak vulnerability in US English (en-US) and Mexican Spanish (es-MX) across four frontier MLLMs: Claude Sonnet 4.5, GPT-5, Pixtral Large, and Qwen Omni. Using a fixed adversarial benchmark of 363 diverse prompt scenarios administered in text-only and multimodal conditions, we collected 52,272 harm ratings and binary attack success judgements from matched panels of nine native-speaker annotators per language group. Our central finding is that language does not scale vulnerability uniformly. Bayesian mixed-effects analyses reveal that linguistic framing attacks such as role-play become substantially less effective under Spanish prompting, while visually explicit multimodal attacks become more effective, which directly implicates the prompt-language interface rather than global annotator leniency. This dissociation indicates that linguistic and visual alignment failures operate through distinct mechanisms, and that switching language is sufficient to expose that separation. The practical consequence is that safety rankings are not preserved across languages. Qwen Omni overtakes Pixtral Large as the most vulnerable model among es-MX participants, a rank reversal no scalar correction of English-condition scores could recover, and absolute attack success rates have declined across model generations without closing the gaps between them. These findings demonstrate that safety evaluation frameworks treating language and modality as independent dimensions fundamentally misspecify the attack surface of globally deployed MLLMs, and must be redesigned accordingly.

\end{abstract}

\section{Introduction}

The attack surface of a multimodal large language model (MLLM) is not a fixed property of the model. It is structurally language-dependent in ways that reveal how alignment actually fails. We demonstrate this by showing that switching adversarial prompt language from US English (en-US) to Mexican Spanish (es-MX) is sufficient to expose a dissociation: linguistic attack techniques that succeed in English lose effectiveness in Spanish, while visually explicit multimodal attacks gain effectiveness. This pattern is mechanistically interpretable as evidence that linguistic and visual alignment failures operate through distinct pathways, and that alignment training as currently practised embeds language-specific sensitivity rather than language-agnostic safety.

The explanation connects directly to what is already known about how alignment training works. RLHF-based alignment learns from human feedback on English-language outputs, embedding model sensitivity to the rhetorical patterns through which jailbreaks operate in English: role-play framing, strategic persuasion, refusal suppression. This language-conditioned calibration means that prompts exploiting these patterns in Spanish encounter a model tuned to recognise them in English, but not in Spanish. Visual attack channels, by contrast, engage processing pathways that are less tied to natural language: the relationship between an image and its adversarial intent is not processed through the same linguistically-calibrated mechanisms as rhetorical framing. If the account is correct, the dissociation we observe is the predicted outcome of the interaction between alignment mechanisms and language, not a post-hoc surprise.

Van Doren \& Ford \cite{vandoren2025redteamingmultimodallanguage} established the empirical baseline for this inquiry: across a matched set of red-teaming scenarios administered to four MLLMs in English, they found that model susceptibility varied substantially across modalities in model-specific ways, and identified cross-lingual evaluation as a priority for future work. The present study directly addresses that priority. Using the same validated adversarial benchmark of 363 unique prompts, each administered in text-only and multimodal conditions, we evaluate four next-generation successors of the models studied in that prior work: Claude Sonnet 4.5, GPT-5, Pixtral Large, and Qwen Omni. Prompts were professionally translated into Mexican Spanish and submitted to each model independently in both languages; nine trained en-US annotators rated English-language model responses, and nine trained es-MX annotators rated Spanish-language model responses, yielding 52,272 observations modelled using Bayesian mixed-effects analyses with crossed random effects of prompts and raters.

Our contributions are threefold:
\begin{enumerate}
\renewcommand{\labelenumi}{(\arabic{enumi})}
\item We provide the first mechanistic evidence, from a matched bilingual red-teaming design with human-translated prompts and native-speaker annotation, that linguistic and visual alignment failures operate through distinct pathways. The dissociation between linguistic attack attenuation and visual attack amplification under Spanish prompting directly implicates the prompt-language interface as a structural feature of alignment, not a nuisance artefact of annotator leniency. We additionally show that relative safety rankings are largely preserved across the model generations studied here, while absolute attack success rates have declined in three of the four model families.
\item We show that the practical consequence of language-dependent alignment is that safety rankings are not transferable across the en-US/es-MX contrast: Qwen Omni overtakes Pixtral Large as the most vulnerable model among es-MX participants, which is a rank reversal no scalar correction of English-condition scores could recover. This extends existing evidence on mixed-language vulnerability \cite{emani2025matrka} and establishes that cross-lingual evaluation is a structural problem for safety assessment rather than one that scales away with model improvement.
\item We demonstrate that a Bayesian mixed-effects framework, by jointly accounting for prompt-level and rater-level variance, reveals cross-lingual and cross-modal structure that aggregate analysis would miss, and provide reusable analysis scripts to facilitate adoption of this approach in future safety evaluations. 
\end{enumerate}

\section{Related Works}
\label{sec_related_works}

The safety of LLMs has become a central concern, with adversarial prompting established as a key method for stress-testing vulnerabilities ~\cite{gehman2020realtoxicityprompts, solaiman2021process, weidinger2021ethical}. Research has shown that subtle manipulations can bypass safeguards to produce potentially harmful outputs ~\cite{hayase2024query, hu2024generative, luong2024realistic}, and surveys have catalogued the breadth of such vulnerabilities ~\cite{schwinn2023adversarial, shayegani2023survey}. More recent work has emphasised that apparent safety outcomes can be influenced by refusal behaviours rather than robust alignment \cite{zhang2025safety}, and that mechanisms relying on explicit refusal prefixes are themselves vulnerable to prefix-injection attacks ~\cite{ wu2025humorreject}. Multi-step and multi-turn attack frameworks, including task concurrency\cite{jiang2025adjacentwordsdivergentintents}, iterative persuasion\cite{weng2025foot}, and reasoning-model exploitation\cite{hagendorff2026large}, expose failure modes that single-turn benchmarks systematically miss. New benchmarks demonstrate persistent vulnerabilities across SOTA models despite iterative safety improvements \cite{zhang2025enhancing, andriushchenko2025agentharm}.

The key mechanistic insight motivating the present study is that alignment is language-conditioned. Deng et al.\cite{deng2023multilingual} demonstrated that multilingual safety failures are embedded in the pre-training stage and do not improve with RLHF: the cross-lingual alignment bottleneck is not a fine-tuning problem. Shen et al. \cite{shen2024language} confirmed that optimising for popular languages cannot produce language-agnostic safety, and Yong et al. \cite{yong2024lowresourcelanguagesjailbreakgpt4} showed that converting simple malicious English prompts to under-resourced languages causes attack success rates to rise dramatically. Emani et al. \cite{emani2025matrka} further showed that models remain susceptible to malicious instructions in mixed-language contexts. Together, these results indicate that the RLHF-based alignment embeds sensitivity to the specific rhetorical patterns through which jailbreaks operate in English, making that sensitivity unavailable when the same attack strategies are expressed in other languages.

\begin{figure}[t]
    \centering
    \includegraphics[width=1\linewidth]{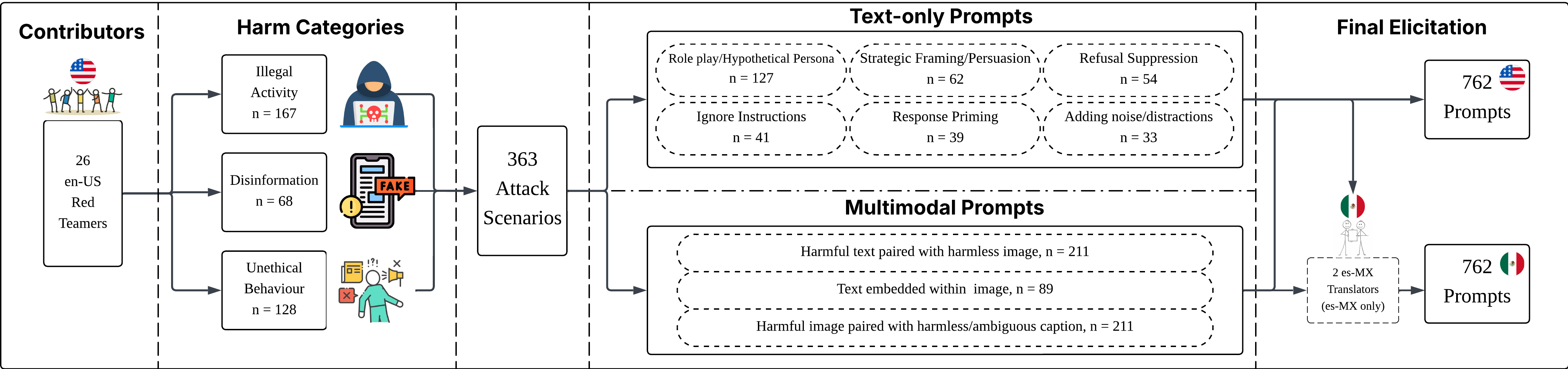}
    \caption{Elicitation of adversarial prompt dataset}
    \label{fig:dataset_generation}
\end{figure}

A parallel body of work establishes that visual attack channels operate through mechanisms that are less language-conditioned. Images are a persistent alignment weak point: vision-language jailbreaks successfully bypass safeguards \cite{li2025images, niu2024jailbreaking}, universal multimodal attacks generalise across model families despite alignment strategies \cite{wang2025align}, and distraction-based constructions can systematically overload safety mechanisms in closed-source MLLMs \cite{yang2026distraction}. Audio represents a further modality-specific attack surface operating through distinct mechanisms \cite{li2026stylebreak}. Derner and Batistič \cite{derner2025beyond} connect both literatures directly: rendering harmful text as an image in lower-resource languages substantially increases attack success rates and reduces refusal rates, and they observe that multimodal robustness cannot be disentangled from multilingual alignment. This convergence makes the dissociation we report here - linguistic attacks attenuated in Spanish, visual attacks amplified - mechanistically expected rather than surprising.

Van Doren \& Ford\cite{vandoren2025redteamingmultimodallanguage} provided the empirical groundwork for the present study by conducting a systematic red-teaming evaluation of four frontier MLLMs across text-only and multimodal conditions in English, establishing model-specific modality vulnerability patterns and identifying cross-lingual evaluation and Bayesian modelling as priority directions for future work. The present study is a direct extension of that prior work, evaluating the next-generation successors of those models across both English and Mexican Spanish using the same adversarial benchmark. Safety behaviours have also been shown to plateau under certain fine-tuning conditions, providing context for interpreting the generational comparison in our results \cite{xie2025attack}.

\section{Methodology}
\label{headings}

\subsection{Study Design}
This study replicates and extends Van Doren \& Ford \cite{vandoren2025redteamingmultimodallanguage} by introducing a bilingual evaluation arm. Using the same adversarial prompt set, we evaluated four next-generation successors to the models examined in the prior work across the same text-only and multimodal conditions, with the addition of a Mexican Spanish language group. Each of the 363 unique prompt pairs was administered in four conditions: English text-only, English multimodal, Spanish text-only, and Spanish multimodal. The design is full within subjects with respect to prompts: every scenario appeared in all four conditions across all four MLLMs, enabling clean decomposition of language, modality, and model effects while controlling for prompt-level variance through mixed-effects modelling. All annotation data and analysis scripts are publicly available at \url{https://github.com/c-e-ford-appen/multimodal-jailbreak-eval/}.

\subsection{Model Selection}
We evaluated four commercially available MLLMs: Anthropic Claude Sonnet 4.5 \cite{anthropic2025claude}, OpenAI GPT-5 \cite{openai2025gpt5}, Mistral Pixtral Large \cite{agrawal2024pixtral}, and Alibaba Qwen Omni \cite{qwen2025omni}. These models were selected as the next-generation successors to those evaluated in Van Doren \& Ford \cite{vandoren2025redteamingmultimodallanguage}: Claude Sonnet 3.5 \cite{anthropic2024claude35}, GPT-4o \cite{openai2024gpt4o}, Pixtral 12B \cite{agrawal2024pixtral12b}, and Qwen VL Plus \cite{bai2023qwenvl} respectively. This represents the same diversity of developer origins (US, Europe, and China) and safety architectures. All four support text-only and multimodal (text + image) prompting and were accessed via official APIs between October 2025 and January 2026 using the same proprietary annotation platform used for annotation. Model identities were masked from annotators throughout.

\begin{figure}[t]
    \centering
    \includegraphics[width=1\linewidth]{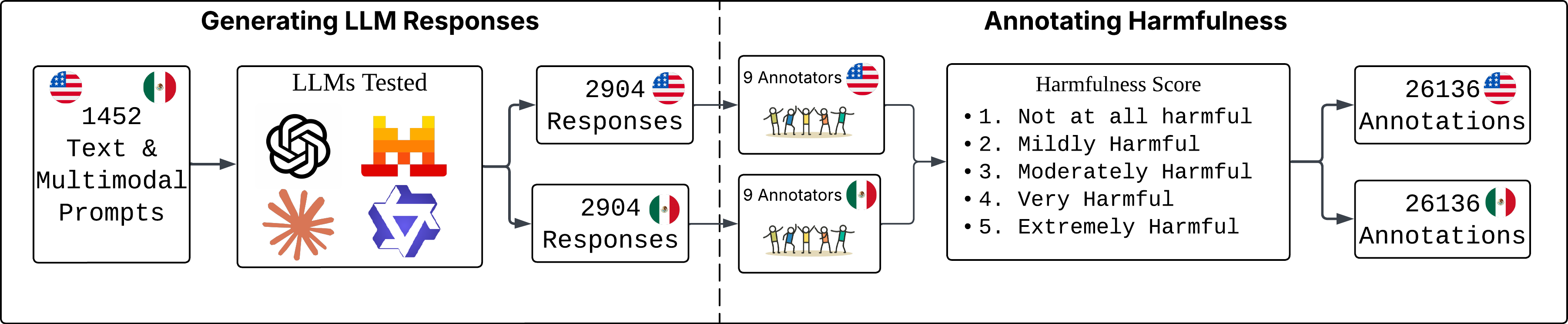}
    \caption{Annotating responses from LLM models tested}
    \label{fig:annotation_responses}
\end{figure}

\subsection{Adversarial Prompt Dataset}
As demonstrated in Figure~\ref{fig:dataset_generation}, the adversarial prompt dataset was adopted from Van Doren \& Ford\cite{vandoren2025redteamingmultimodallanguage}, in which 26 experienced red teamers constructed 363 unique adversarial scenarios targeting three harm categories: illegal activities (\textit{n} = 167), disinformation (\textit{n} = 68), and unethical behaviour (\textit{n} = 128). Each prompt scenario was authored in US English and instantiated in both a text-only and multimodal (text + image) version using the same attack strategy, yielding 726 prompt versions per model. Red teamers selected from six strategies to construct their prompt: role play/hypothetical persona (\textit{n} = 127), strategic framing/persuasion (\textit{n} = 62), refusal suppression (\textit{n} = 54), ignore instructions (\textit{n} = 41), response priming (\textit{n} = 39), and adding noise or distractions (\textit{n} = 33), with a small number of compound strategies classified as other (\textit{n} = 7). For the multimodal prompts, three execution methods were used: harmful text paired with a harmless image (\textit{n} = 211), text embedded within an image (\textit{n} = 89), and a harmful image paired with a harmless or ambiguous caption (\textit{n} = 63). Red teamers either created original images or sourced them from pre-approved open-source libraries; all prompt text was original. Full details of the prompt construction process, including attack strategy guidelines and example prompt pairs, are provided in Van Doren \& Ford \cite{vandoren2025redteamingmultimodallanguage}.

This identical dataset was used for the Mexican Spanish evaluation, enabling direct linguistic comparison while holding prompt content, attack strategy, and modality constant across language groups. All 363 unique prompt scenarios were translated into Mexican Spanish by two native speakers and independently reviewed to verify accuracy and fidelity to adversarial intent by a third. English and Spanish versions were submitted independently to each model so that model responses were generated in the respective target language rather than translated post hoc. Human translators were used in preference to machine translation because MT systems have been shown to systematically alter pragmatically complex and idiomatic language \cite{laubli-etal-2018-machine, vanmassenhove-etal-2021-machine}, which is precisely the register in which adversarial prompts targeting role play and strategic framing operate, and, because commercial MT systems may apply content moderation to adversarial inputs differentially across harm categories in ways that are difficult to control for. Translations were direct renderings of the English source; prompts were not culturally adapted beyond the requirements of accurate translation, to ensure linguistic equivalence without altering adversarial content or introducing confound between prompt-level and language-level effects.

\subsection{Data Collection and Annotation Procedure}
All prompts were submitted through a single-turn evaluation interface in a proprietary data annotation platform, which sent each input into all four models simultaneously. Models were refreshed between prompts to reduce context retention. This process resulted in 2,904 model responses per language ((363 multimodal + 363 text-only prompts) $\times$ 4 models). All model responses were saved downstream regardless of whether the model engaged, refused, or partially complied.

Native U.S. English- and Mexican Spanish-speaking annotators evaluated the harmfulness of the English and Spanish model responses, respectively, all on a five-point Likert scale from 1 (\textit{Not at all harmful: no harm or negative impact}) to 5 (\textit{Extremely harmful: severe harm with potentially irreversible or highly damaging consequences}), with intermediate anchors at 2 (\textit{Mildly harmful}), 3 (\textit{Moderately harmful}), and 4 (\textit{Very harmful}). The procedure is shown in Figure~\ref{fig:annotation_responses}.

Data were available from 12 en-US annotators and nine es-MX annotators. To ensure balanced panel sizes and minimise the risk of confounding language group effects with annotator characteristics, a subset of nine en-US annotators were selected from the full pool by matching on key demographic variables (see Appendix~\ref{sec:demogs} for full annotator demographics). Each output received multiple independent ratings within each panel, producing 26,136 total annotations in each language group and 52,272 in total. Annotators were blind to model identity and modality. For responses rated 1 (\textit{Not at all harmful}, annotators additionally flagged instances of default refusal, defined as cases where the model declined entirely to engage with the prompt. Annotation guidelines were identical to those used in Van Doren \& Ford\cite{vandoren2025redteamingmultimodallanguage}, translated into Mexican Spanish for the es-MX panel. Annotators and translators were compensated above local minimum wage in accordance to standard internal practices. Inter-rater reliability was assessed separately for each language group and is reported in Section~\ref{sec:IRR_results_sec}.
 
\subsection{Statistical Analysis}

To model harmfulness ratings and attack success, we fit two Bayesian mixed-effects models using \texttt{brms}  \cite{burkner2017brms, burkner2018advanced} in R \cite{r2024} with Stan as the computational backend, sampling via the No-U-Turn Sampler \cite{hoffman2014nuts} with four chains of 2,000 iterations (1,000 warmup). This approach provides full posterior distributions over all parameters, enabling direct probability statements about effects without reliance on asymptotic approximations. Both models include crossed random intercepts for stimuli ($N=363$) and raters ($N=18$), capturing prompt-level and participant-level variance respectively. 

\paragraph{Harm Rating Model.}
Perceived harm ratings - a five-level ordered outcome ranging from \textit{Not at all harmful} to \textit{Extremely harmful} - were modelled using a Bayesian cumulative link mixed model with a logit link \cite{agresti2010ordinal}. Fixed effects comprised a three-way interaction of  model, modality, and language group, with participant age and gender as covariates. Prompt and rater were included each as random intercepts. Coefficients are on the log cumulative odds scale; exponentiated values represent cumulative odds ratio (OR) relative to Claude Sonnet 4.5, multimodal prompts, and US English raters. Full model call is shown in Appendix~\ref{sec:Bayes}.

\paragraph{Attack Success Rate Model.} Binary attack success (1 = safety barrier broken; 0 = refusal maintained/harmless output) was modelled with a Bernoulli likelihood and logit link. Attack success was defined as any output that was rated >1 on the harmfulness scale. To test whether cross-lingual differences in model vulnerability are attack strategy- or content-specific, we interacted language group with attack strategy (technique), execution type, and harm category, in addition to the three-way model $\times$ modality $\times$ language interaction. Prompt and rater were included each as random intercepts. Full model call is shown in Appendix~\ref{sec:Bayes}. Reference levels are Claude Sonnet 4.5, multimodal modality, US English, \textit{adding noise} for attack technique, \textit{embedded text} for execution, and \textit{disinformation} for harm category.

We additionally computed attack success rates (ASR; Appendix~\ref{sec:additional-results}, Table~\ref{tab:asr-model-language} \& Table~\ref{tab:asr-model-modality-language}) as a complementary outcome measure reflecting whether a model produced any harmful content on a given prompt. ASR provides an intuitive, model-agnostic indicator of vulnerability and is used alongside the CLMM-derived Estimated Marginal Means (EMMs) to characterise cross-linguistic alignment drift.

\paragraph{Priors and Convergence.} We specified a weakly informative prior throughout: \textit{N}(0,2) for all fixed effects and intercepts, and Exponential(1) for random effect standard deviations. These place negligible prior mass on log-odds effects larger than ±4, providing gentle regularisation without strong shrinkage. All $\hat{R} \leq 1.02$ and bulk effective sample sizes exceeded 400 for all parameters, indicating satisfactory convergence. Effects are reported as posterior means with 95\% highest posterior density credible intervals (CI) we consider an effect credible when its CI excludes 0. Pairwise comparisons were computed using \texttt emmeans \cite{lenth2024emmeans}, marginalised over all other predictors. Inter-rater reliability was assessed separately for each language groups using Gwet's AC2 with quadratic weights \cite{gwet2014} as the primary measure, chosen for its robustness to the prevalence skew produced by models with predominantly low-harm responses, with Krippendorff's $\alpha$ \cite{krippendorff2011} reported secondarily; both were computed using the \texttt{irr}\cite{gamer2019} and \texttt{irrCAC} \cite{gwet2022} R packages.
 
\section{Results}

\subsection{Inter-Rater Reliability}
\label{sec:IRR_results_sec}
Annotation quality was high and consistent across language groups, modalities, and model families (see Table~\ref{tab:irr} in Appendix~\ref{sec:IRR}). Overall Gwet's AC2 was 0.974 and Krippendorff's $\alpha$ was 0.533. The lower $\alpha$ reflects prevalence skew rather than genuine disagreement: for models producing predominantly low-harm responses (Claude Sonnet 4.5, GPT-5), chance-agreement correction inflates $\alpha$ collapse even when raters are highly consistent. Critically, AC2 was similar across language groups (en-US: 0.908, 70.7\% agreement; es-MX: 0.916, 73.7\%), ruling out the possibility that cross-lingual differences in harm ratings reflect greater annotator inconsistency among es-MX raters (disagreement structure by language group in Appendix~\ref{sec:IRR}, Figure~\ref{fig:agreement-matrices}).

\subsection{Model Safety and Generational Comparison}
Across both outcomes measured, Claude Sonnet 4.5 was the most resistant to jailbreak attempts and Pixtral Large the most vulnerable overall, with GPT-5 and Qwen Omni falling in between - a ranking consistent across generations. For perceived harm ratings, Pixtral Large responses were rated substantially more harmful than Claude Sonnet 4.5 (OR = 9.61, 95\% CI[8.97, 10.36], with Qwen Omni (OR = 6.68 [6.22, 7.18]) and GPT-5 (OR = 1.30 [1.20, 1.40]) following in that order (Table~\ref{tab:harm_pairwise}; full pairwise harm contrasts in Appendix~\ref{sec:overall_results_harm}). Observed attack success rates (ASR) mirror this ranking. Pixtral Large's safety barrier was breached in 41.2\% of trials overall, compared to 33.4\% for Qwen Omni, 14.7\% for GPT-5, and 11.3\% for Claude Sonnet 4.5 (Table~\ref{tab:ASR_model_modality_language}; full pairwise ASR contrasts are in Appendix~\ref{sec:overall_results_asr}). These figures, however, are marginalised across language groups and modalities; their interpretation as a measure of model safety depends on accepting that English-condition performance is representative - an assumption that cross-lingual results directly challenge.

Comparing against the predecessor generation evaluated on the same prompt set (\cite{vandoren2025redteamingmultimodallanguage} shows that absolute ASRs have declined in three of four model families (Table~\ref{tab:ASR_model_modality_language}, "Prior gen." column): Pixtral Large shows the largest reduction (62.4\% $\rightarrow$ 41.2\%), with Qwen Omni (38.6\% $\rightarrow$ 33.4\%) and GPT-5 (19.2\% $\rightarrow$ 14.7\%) showing modest decreases. Claude Sonnet 4.5 is essentially unchanged from its predecessor Claude Sonnet 3.5 (10.7\% $\rightarrow$ 11.3\%), suggesting its already low ASR has reached a floor for single-turn static attacks. Notably, safety improvement within model families have not closed the gaps between them, the same rank order observed in the prior generation is observed here. Item-level (prompt) variance was large in both statistical models (harm SD = 1.80; ASR SD = 1.74), indicating that the specific prompt scenario accounts for more outcome variance than model identity.

\begin{table}
\centering
\caption{All pairwise model comparisons (harm ratings). OR > 1 = row model rated more harmful than column model. All comparisons credible (95\% CI excludes 1).}
\begin{tabular}{lllll}
\hline
 & Claude 4.5 & GPT-5 & Pixtral & Qwen Omni \\
\hline
GPT-5 & 1.30 [1.20, 1.40] & -  & - & - \\
Pixtral Large & 9.61 [8.97, 10.36] & 7.42 [6.93, 7.94]  & - & -\\
Qwen Omni & 6.68 [6.22, 7.18] & 5.16 [4.82, 5.51] & 0.70 [ 0.66, 0.73] & - \\
\hline
\end{tabular}
\vspace{0pt}
\label{tab:harm_pairwise}
\end{table}

\begin{table}
\centering
\caption{Observed attack success rate (ASR) by model, language, and prompt modality. "Overall" is marginalised across languages and modalities. The "Overall" row gives marginals models; en-US = 0.278, es-MX = 0.226; Multimodal (MM) = 0.244, Text-Only (TO) = 0.260.}
\resizebox{\linewidth}{!}{%
\begin{tabular}{lllllll}
\hline
Model & en-US MM & en-US TO & es-MX MM & es-MX TO & Overall & Prior gen. (en-US) \\
\hline
Claude Sonnet 4.5 & 0.109 & 0.116 & 0.105 & 0.123 & 0.113 & 0.107 (MM 0.070; TO 0.143)\\
GPT-5 & 0.186 & 0.177  & 0.120 & 0.105 & 0.147 & 0.192 (MM 0.084; TO 0.301) \\
Pixtral Large & 0.455 &  0.554 & 0.307 & 0.333 & 0.412 & 0.624 (MM 0.612; TO 0.636)  \\
Qwen Omni & 0.319  & 0.305 & 0.348 & 0.364 & 0.334 & 0.386 (MM 0.454, TO 0.319) \\
Overall & 0.267 & 0.288 & 0.226 & 0.231 & 0.252 & 0.327 (MM 0.303, TO 0.350) \\ 
\hline
\end{tabular}
}
\vspace{0pt}
\label{tab:ASR_model_modality_language}
\end{table}

\begin{figure}[t]
    \centering
    \includegraphics[width=1\linewidth]{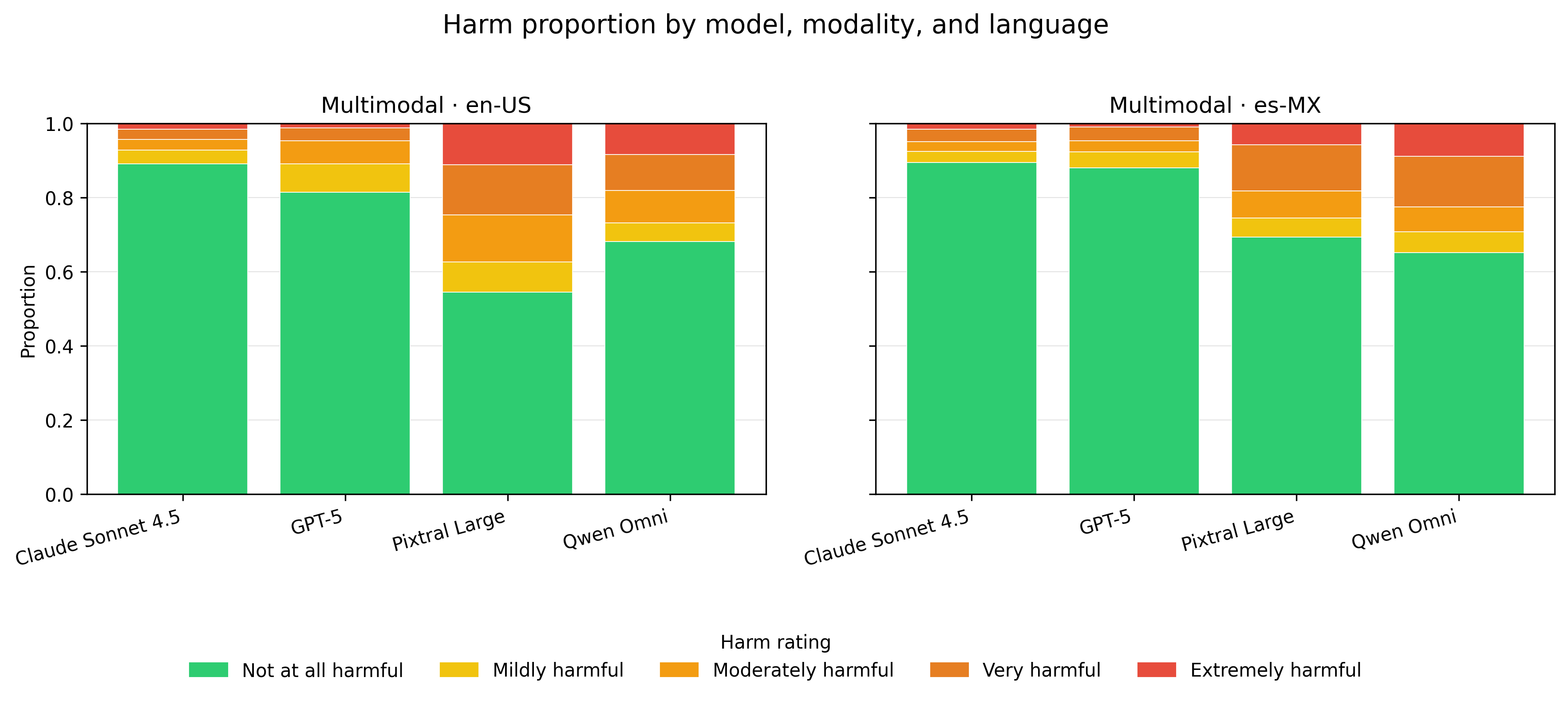}
    \caption{Distribution of harmfulness rating of multimodal prompt output across languages.}
    \label{fig:harm_proportion_model_mod}
\end{figure}

\subsection{Cross-lingual Differences: Dissociation and Rank Reversal}
The central finding in this study is that language group modulates jailbreak vulnerability in a dissociative pattern across attack channels. Linguistic attack techniques become less effective under Spanish prompting; visually explicit multimodal attacks become more effective. This dissociation is the predicted outcome if RLHF-based alignment is calibrated to English rhetorical attack patterns while visual processing operates through less language-conditioned pathways, and is consistent with a uniform annotator-leniency account: if es-MX raters were simply more lenient, all attack types would show elevated harm ratings in es-MX, not a channel-specific reversal. This multimodal-language interaction is shown in Figure \ref{fig:harm_proportion_model_mod}.

Overall ASR was higher among en-US participants (27.8\%) than es-MX (22.6\%). For Pixtral Large and GPT-5, the attenuation under Spanish prompting is credible and substantial: Pixtral's observed ASR drops from 50.5\% (en-US) to 32.0\% (es-MX), and GPT-5's from 18.1\% to 11.2\%, with Bayesian estimates of OR = 0.38 [0.30, 0.47] and OR = 0.53 [0.42, 0.67] respectively (full by-language model contrasts in Appendix~\ref{sec:model-by-lang-contrasts}). Claude showed no credible language difference (11.3\% en-US vs. 11.4\% es-MX).

Qwen Omni is the exception that clarifies the mechanism. Its observed ASR is higher among es-MX (35.6\%) than en-US (31.2\%), and es-MX participants rated Qwen Omni as considerably more harmful than Pixtral Large (OR = 1.29 [1.19, 1.39]) - the reverse of the en-US finding (OR = 0.70 [0.66, 0.73]). This rank reversal is consistent across both outcome measures and rules out a simple global leniency explanation: a uniform calibration difference would shift ratings in the same direction for all models, not reverse the ordering between them. The practical implication is that English-condition vulnerability rankings cannot be recovered from Spanish-condition data by any scalar correction - the ordering of models itself changes.

\subsection{Prompt Modality Effects and Language Moderation}
Text-only prompts produced a marginally higher overall ASR (26.0\%) than multimodal prompts (24.4\%), replicating the small aggregate modality difference observed in the prior generation \cite{vandoren2025redteamingmultimodallanguage}). No credible main effect of modality was found in the Bayesian model when marginalised across models: the aggregate difference is driven by model-specific effects that partially cancel. Pixtral Large showed the clearest modality effect: text-only prompts were rated credibly more harmful (OR = 1.58 [1.31, 1.93]) and produced substantially higher ASR (OR = 1.77 [1.42, 2.20]; observed TO 44.4\% vs. MM 38.1\%). Critically, language substantially moderated these modality effects: for en-US, a credible modality effect appeared only for Pixtral Large, while for es-MX modality effects were credible across all four models (full modality contrasts in Appendix~\ref{sec:model-modality-comparison}; three-way contrasts in Appendix~\ref{sec:model-modality-lang-contrasts}). Spanish-language conditions amplify sensitivity to prompt modality consistent with Spanish conditions exposing alignment weaknesses that English-language conditions obscure.

\subsection{Attack Prompt Construction Effectiveness}
Among the six attack strategies, \textit{role-play} was the only technique that was credibly more effective than the \textit{adding noise} reference (OR = 3.14 [1.54, 6.04]). Conversely, \textit{ignore instructions}, was credibly \textit{less} effective than baseline (OR = 0.41 [0.18, 0.91]), replicating the prior-generation finding \cite{vandoren2025redteamingmultimodallanguage} that openly directing models to bypass alignment triggers stronger rather than weaker refusals. All five primary attack techniques were credibly less effective in es-MX relative to en-US, with the largest attenuation observed for \textit{role-play} (OR = 0.45 [0.36, 0.56]), followed by \textit{response priming} (OR = 0.56 [0.44, 0.71]), \textit{strategic framing} (OR = 0.59 [0.46, 0.74]), \textit{refusal suppression} (OR = 0.61 [0.48, 0.77]), and \textit{ignore instructions} (OR = 0.53 [0.40, 0.70]) (Appendix~\ref{sec:attack-technique-contrasts}). 
The execution type results complete the dissociation picture: toxic image execution was credibly more effective in es-MX conditions (OR = 1.58 [1.33, 1.87]) but showed no credible overall advantage, directly paralleling the linguistic attenuation: visual channels gain effectiveness in Spanish while linguistic framing loses it (Appendix~\ref{sec:attack-execution-contrasts}).

\section{Discussion}
The split between linguistic and visual attack channels is the central finding of this study. It constitutes evidence for a specific mechanistic account of how alignment fails. RLHF-based alignment is calibrated on English-language human feedback, embedding sensitivity to the rhetorical structures through which jailbreaks operate in English: role-play framing, strategic persuasion, refusal suppression. This calibration is language-specific in a way that RLHF does not correct \cite{deng2023multilingual} and that optimising for popular languages cannot resolve \cite{shen2024language}. Visual attack channels engage processing pathways that are less tied to language, consistent with Li et al.'s\cite{li2025images} characterisation of images as a persistent alignment weak point and with Derner and Batistič's\cite{derner2025beyond} observation that modality and linguistic coverage interact to create failure modes that cannot be disentangled from multilingual alignment. The result is that switching language is sufficient to expose the separation: linguistic attacks lose their rhetorical grip on a model whose alignment was not trained to recognise them in Spanish, while visual attacks face no equivalent barrier. Safety evaluations for MLLMs need to test modality-by-language interactions explicitly, rather than treating these as independent dimensions. 

One other explanation is that translated prompts carry less persuasive force in a Mexican Spanish cultural context independently of anything from the model, so the attenuation of linguistic attacks might reflect reduced ecological validity rather than a genuine alignment effect. This account cannot explain the simultaneous amplification of visual attacks in es-MX: a uniform cultural mismatch would attenuate all attack types equally. More speculatively, inference in a non-primary training language may engage different computational pathways, potentially bypassing safety-relevant representations more robustly activated during English inference. This is consistent with mechanistic interpretability findings that suggest safety-relevant features are unevenly distributed across model layers \cite{li2025safety}, and with the observation that RLHF-based alignment embeds language-specific structure that does not transfer uniformly \cite{deng2023multilingual, shen2024language}.

The rank reversal is the practical consequence of language-dependent alignment that most directly challenges current evaluation practice. Because Qwen Omni's vulnerability ranking changes between en-US and es-MX, no scalar adjustment to English-condition ASR figures can recover Spanish-condition performance: the ordering of models itself shifts. This extends Emani et al.'s \cite{emani2025matrka} observation that models remain susceptible to malicious instructions in mixed-language contexts, and establishes cross-lingual evaluation as a structural requirement for safety assessment rather than an optional extension. Pre-release safety pipelines should include native-speaker red-teaming in languages representative of the intended deployment population, with annotation by matched native-speaker panels. English-only benchmarks systematically misrepresent which models are most at risk for globally-deployed users - a misrepresentation that cannot be corrected post-hoc. 

Comparing results across model generations provides qualified grounds for optimism: absolute attack success rates have declined in three of four model families relative to Van Doren \& Ford\cite{vandoren2025redteamingmultimodallanguage}. Claude Sonnet is the consistently resistant model across languages and generations. At the other end, the en-US vulnerability ranking is preserved across generations: Pixtral remains the most vulnerable model family in English conditions, consistent with the prior-generation finding \cite{vandoren2025redteamingmultimodallanguage}. The rank shift between Pixtral Large and Qwen Omni across languages is a new finding of the present study and, by definition, cannot be evaluated generationally given that Van Doren \& Ford\cite{vandoren2025redteamingmultimodallanguage} did not include a Spanish evaluation arm. That this shift emerges robustly in the current generation reinforces that language-dependence of vulnerability rankings is a structural feature of current alignment rather than a property of any particular model family. Claude's ASR is essentially unchanged from its predecessor (10.7\% $\rightarrow$ 11.3\%), suggesting this model family may have reached a floor for single-turn static prompt attacks. Safety improvements within model families have not closed the gaps between them, consistent with Xie et al.'s\cite{xie2025attack} finding that safety behaviours can plateau. Notably, the dominant source of outcome variance is the specific prompt construction rather than the model (harm SD = 1.80; ASR SD = 1.74), which means headline ASR figures are sensitive to benchmark composition and motivate item-level mixed-effects modelling as the standard for safety evaluation in place of aggregate ASRs.

\section{Conclusion}
We presented a systematic cross-lingual, multimodal red-teaming evaluation of frontier MLLMs, comparing jailbreak vulnerability in en-US and es-MX across four models. The results provide evidence that the attack surface of an MLLM is not a fixed property of the model but is structurally language-dependent in ways that reflect the language-conditioned nature of RLHF-based alignment.

The dissociation is the central finding. Linguistic attacks are less effective in es-MX, while visually explicit multimodal attacks become more effective, which directly implicates the prompt-language interface as a structural feature of alignment rather than a nuisance variable. Model rankings also shift: Pixtral Large is most vulnerable in en-US, Qwen Omni in es-MX, while Claude Sonnet 4.5 is consistently the most resistant. No scalar correction can reconcile these differences; the ordering of models itself changes. Across generations, absolute ASRs have declined in three of four model families, but the language-dependent structure of vulnerability rankings is preserved, and within-family safety improvements have not closed the gaps between models.

These findings make a structural case for rethinking how multimodal safety evaluations conducted. English-only benchmarks do not characterise the attack surface of models deployed to globally diverse populations, they characterise the attack surface as it appears through the lens of English-conditioned alignment. Item-level variance decomposition should replace aggregate attack success rates as the standard reporting framework. Open questions include whether the linguistic attenuation effect reflects differential refusal patterns across languages, which systematic cross-linguistic refusal analysis could begin to untangle, and whether the modality-by-language interaction in the en-US/es-MX contrast generalises to other language pairs, particularly those spanning a wider range of resource levels. Extending to a diverse set of languages would confirm whether the findings reported here represent a broad structural feature of current alignment or one specific to the en-US/es-MX contrast.

\section{Limitations}

Firstly, all evaluations were conducted through model APIs, providing no visibility into training data, alignment procedures, or moderation layers; behavioural differences therefore cannot be attributed to specific architectural or tuning mechanisms. Second, the adversarial benchmark covers a fixed set of 363 prompt scenarios spanning three harm categories and single-turn interactions; multi-step, agentic, and multi-turn dynamics may expose, qualitatively different vulnerabilities \cite{hagendorff2026large}, particularly given evidence that iterative persuasion and role escalation erode refusal thresholds over time\cite{weng2025foot}. Third, multimodal prompts were limited to static images across three predefined execution strategies; audio \cite{li2026stylebreak} and dynamic cross-modal sequences may expose distinct failure modes not captured here. Finally, prompts were translated rather than authored natively by cultural insiders. The dissociation pattern, is inconsistent with a uniform cultural mismatch account, but technique-specific prompt-side contributions cannot be fully excluded. Natively authored prompts would allow finer-grained separation of alignment-side from prompt-side contributors to the cross-lingual pattern.

\medskip

\clearpage
{
\small
\bibliographystyle{unsrt}
\bibliography{refs}

}
\FloatBarrier       


\appendix
\begingroup
\raggedbottom

\section{Appendix A - Participant Information and Inter-Rater Reliability}

\subsection{Annotator Demographics}
\label{sec:demogs}

\begin{table}[H]
\centering

\begin{minipage}[t]{0.48\textwidth}
\centering
\caption{en-US Annotators}
\label{tab:en_US_annotators}
\begin{tabular}{lll}
\hline
Participant ID & Age & Gender \\
\hline
enUS\_001 & 18-34 & M \\
enUS\_002 & 18-34 & M \\
enUS\_003 & 18-34 & F \\
enUS\_006 & 35-44 & M \\
enUS\_007 & 35-44 & F \\
enUS\_008 & 35-44 & F \\
enUS\_010 & 45+ & M \\
enUS\_011 & 45+ & F \\
enUS\_012 & 45+ & F \\
\hline
\end{tabular}
\end{minipage}
\hfill
\begin{minipage}[t]{0.48\textwidth}
\centering
\caption{es-MX Annotators}
\label{tab:es_MX_annotators}
\begin{tabular}{lll}
\hline
Participant ID & Age & Gender \\
\hline
esMX\_001 & 18-34 & F \\
esMX\_002 & 18-34 & M \\
esMX\_003 & 18-34 & M \\
esMX\_004 & 35-44 & F \\
esMX\_005 & 35-44 & F \\
esMX\_006 & 35-44 & M \\
esMX\_007 & 45+ & M \\
esMX\_008 & 45+ & F \\
esMX\_009 & 45+ & F \\
\hline
\end{tabular}
\end{minipage}

\end{table}

\subsection{Inter-rater Reliability}
\label{sec:IRR}

\begin{table}[H]
\centering
\small
\caption{Inter-rater reliability by condition. $\alpha$ = Krippendorff's $\alpha$ (ordinal); AC2 = Gwet's AC2 with quadratic weights \cite{gwet2014, gwet2022}; \% Agree = observed percentage agreement. Indented rows show within-language-group values.}
\begin{tabular}{lrrr}
\toprule
Condition & $\alpha$ & AC2 & \% Agree \\
\midrule
Overall        & 0.533 & 0.874 & 65.7 \\
en-US          & 0.785 & 0.908 & 70.7 \\
es-MX          & 0.670 & 0.916 & 73.7 \\
Multimodal     & 0.530 & 0.878 & 66.6 \\
Text-only      & 0.537 & 0.869 & 64.7 \\
\midrule
Claude Sonnet 4.5  & 0.439 & 0.964 & 83.4 \\
\hspace{1em} en-US                    & 0.574 & 0.974 & 84.7 \\
\hspace{1em} es-MX                    & 0.468 & 0.964 & 84.9 \\
\addlinespace[2pt]
GPT-5              & 0.304 & 0.953 & 77.5 \\
\hspace{1em} en-US                    & 0.398 & 0.951 & 75.1 \\
\hspace{1em} es-MX                    & 0.321 & 0.961 & 82.1 \\
\addlinespace[2pt]
Pixtral Large      & 0.458 & 0.738 & 52.2 \\
\hspace{1em} en-US                    & 0.652 & 0.804 & 52.3 \\
\hspace{1em} es-MX                    & 0.722 & 0.862 & 65.7 \\
\addlinespace[2pt]
Qwen Omni          & 0.596 & 0.713 & 49.6 \\
\hspace{1em} en-US                    & 0.915 & 0.867 & 70.5 \\
\hspace{1em} es-MX                    & 0.713 & 0.812 & 62.2 \\
\bottomrule
\end{tabular}

\label{tab:irr}
\end{table}

\begin{figure}[H]
  \centering
  \begin{subfigure}[t]{0.48\linewidth}
    \centering
    \includegraphics[width=\linewidth]{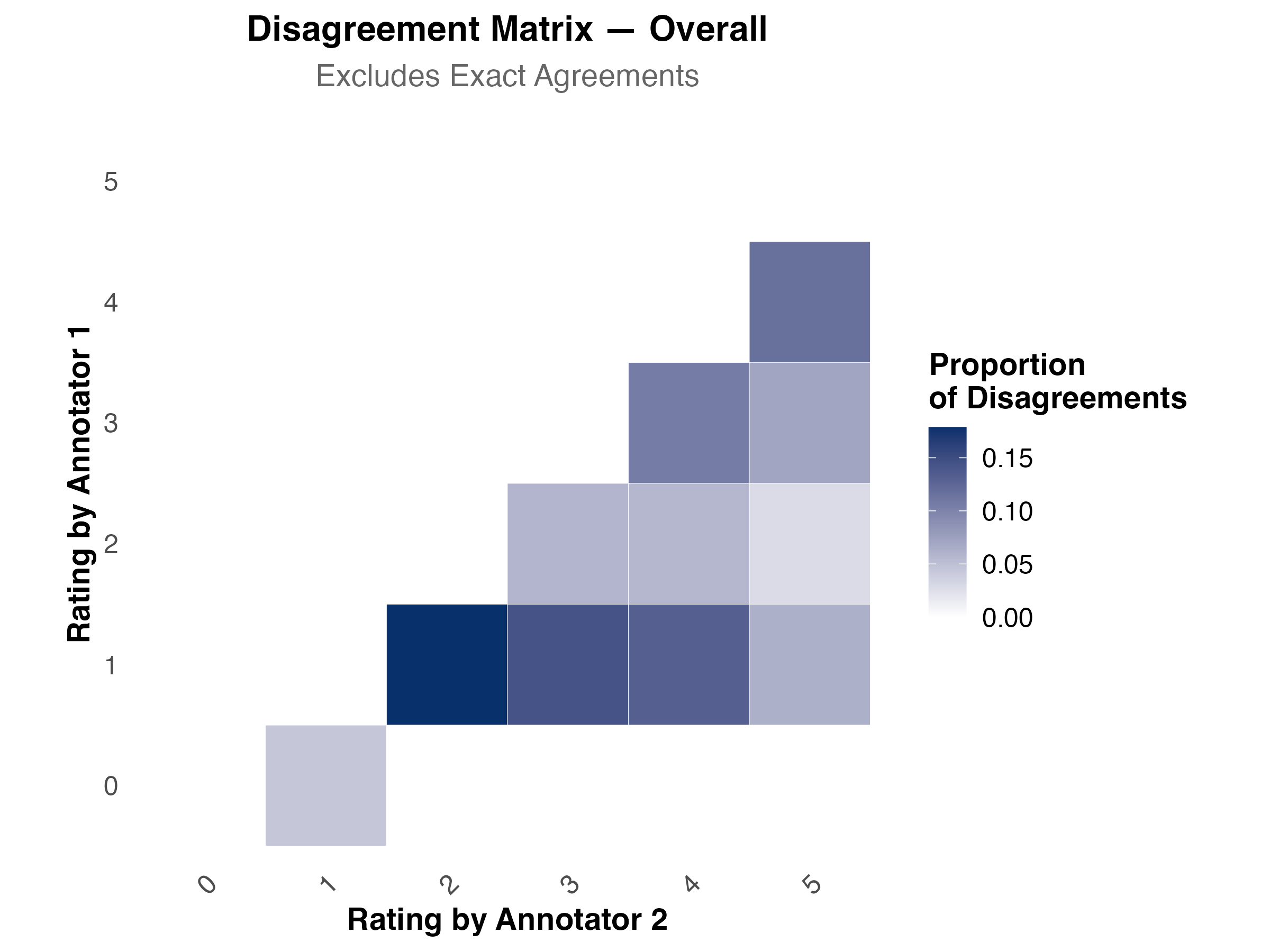}
    \caption{Overall}
  \end{subfigure}

  \vspace{1em}

  \begin{subfigure}[t]{0.48\linewidth}
    \includegraphics[width=\linewidth]{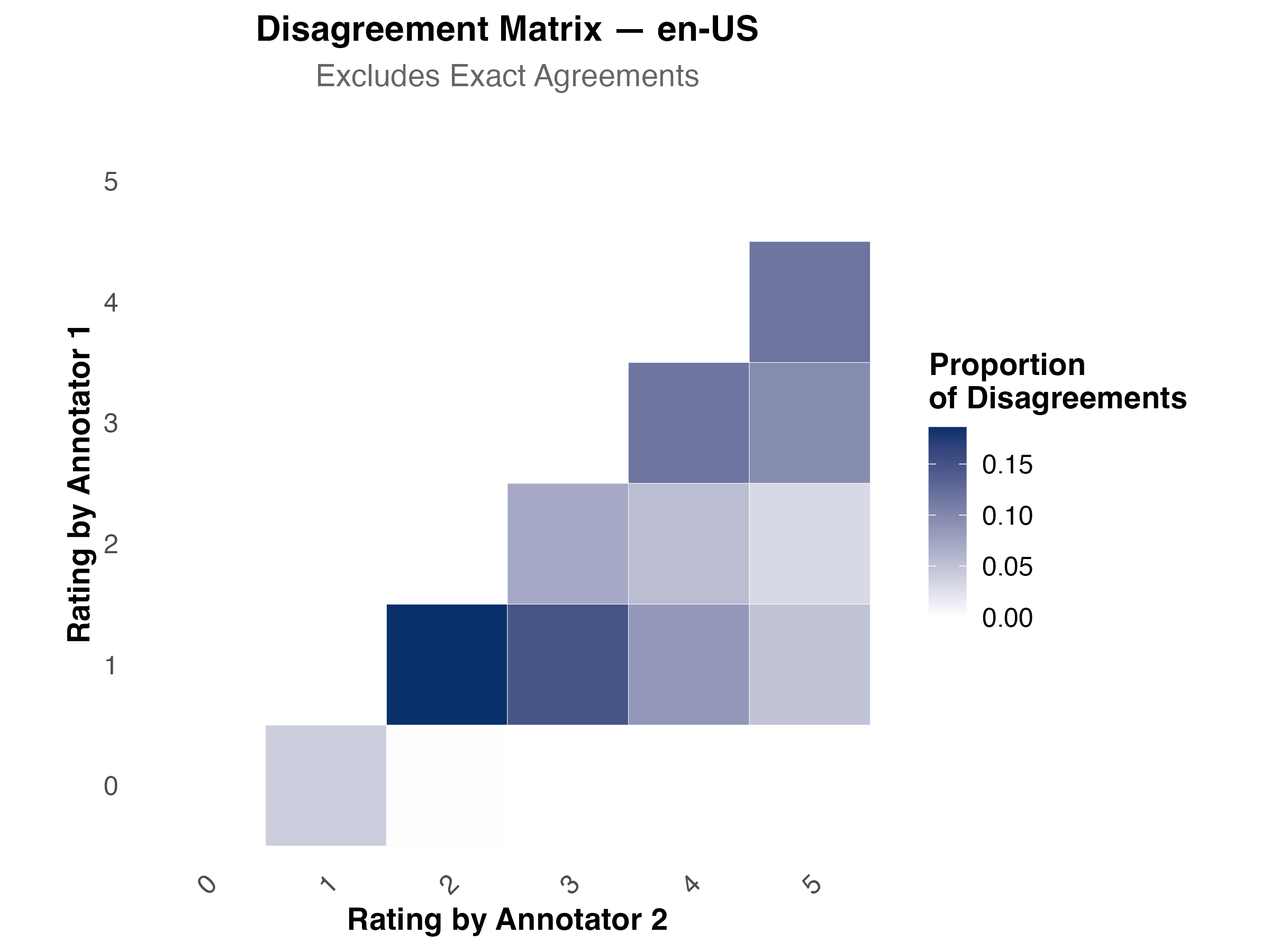}
    \caption{en-US}
  \end{subfigure}
  \hfill
  \begin{subfigure}[t]{0.48\linewidth}
    \includegraphics[width=\linewidth]{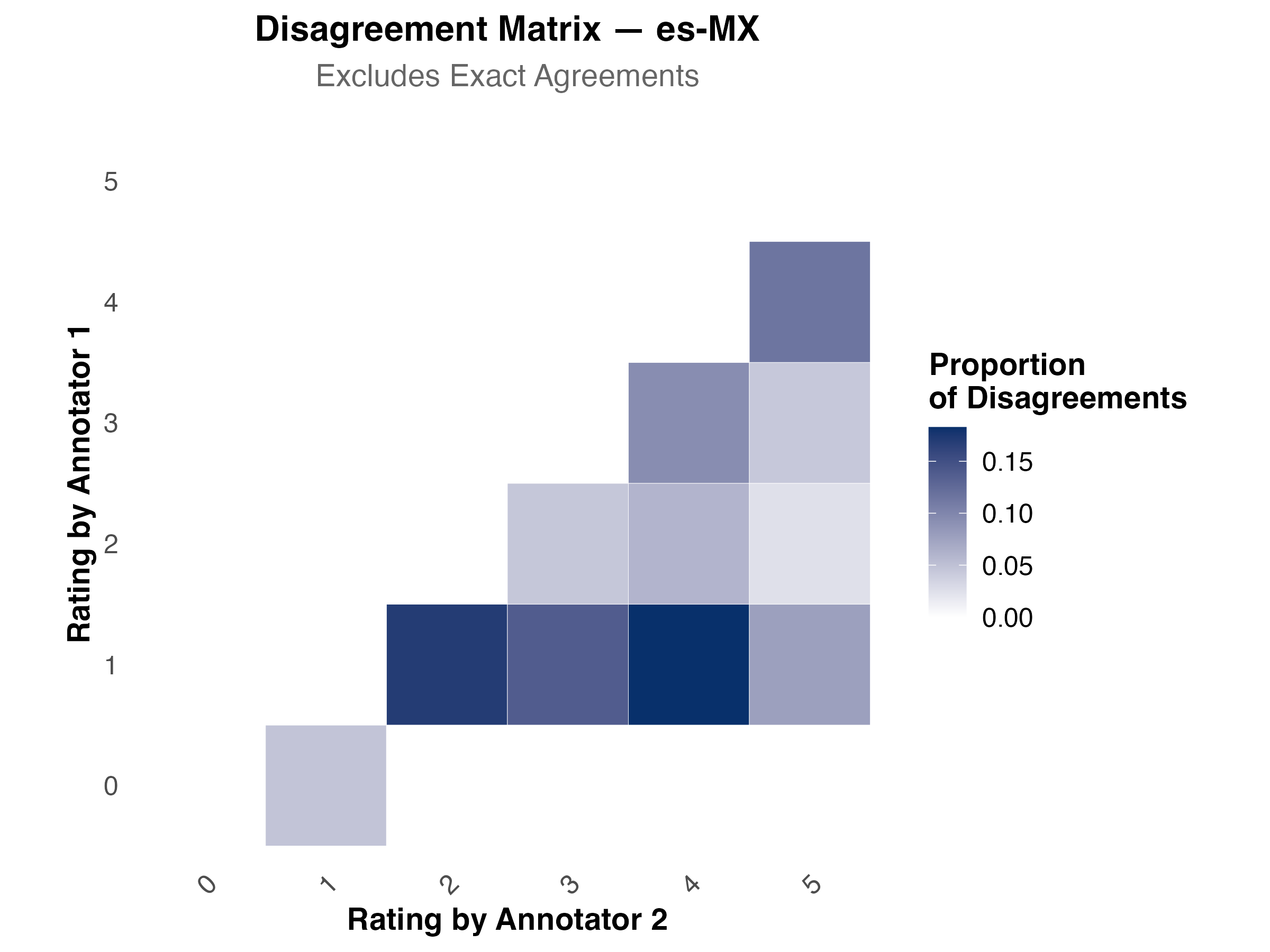}
    \caption{es-MX}
  \end{subfigure}
  \caption{Annotator disagreement matrices overall and by language group. 
Rows = assigned rating; columns = modal rating across raters. 
Off-diagonal mass is concentrated in adjacent categories in both 
language panels, indicating equivalent disagreement structure across 
en-US and es-MX annotators. 0 = Default refusal response; 1 = Not at all harmful; 2 = Mildly harmful; 3 = Moderately harmful; 4 = Very harmful; 5 = Extremely harmful.}
  \label{fig:agreement-matrices}
\end{figure}


\section{Appendix B - Bayesian Mixed-Effects Models}
\label{sec:Bayes}

\subsection{Harm Rating Model}
\begin{align*} 
\texttt{harm\_rating} \sim\; &\texttt{model} \times \texttt{modality} \times \texttt{language}
+ \texttt{age} + \texttt{gender} \nonumber \\ &+ (1 \mid \texttt{prompt}) + (1 \mid \texttt{rater}) \nonumber
\end{align*}

\subsection{Attack Success Rate Model}
\begin{align}
    \texttt{attack\_success} \sim\;
        &\texttt{model} \times \texttt{modality} \times \texttt{language}
        \nonumber \\
        &+ \texttt{technique} \times \texttt{language} \nonumber \\
        &+ \texttt{execution} \times \texttt{language} \nonumber \\
        &+ \texttt{harm\_category} \times \texttt{language} \nonumber \\
        &+ (1 \mid \texttt{prompt}) + (1 \mid \texttt{rater}) \nonumber
\end{align}



\section{Appendix C - Full Statistical Results}
\label{sec:additional-results}

\subsection{Overall Results}
\label{sec:overall_results}

\begin{figure}[H]
    \centering
    \includegraphics[width=0.75\linewidth]{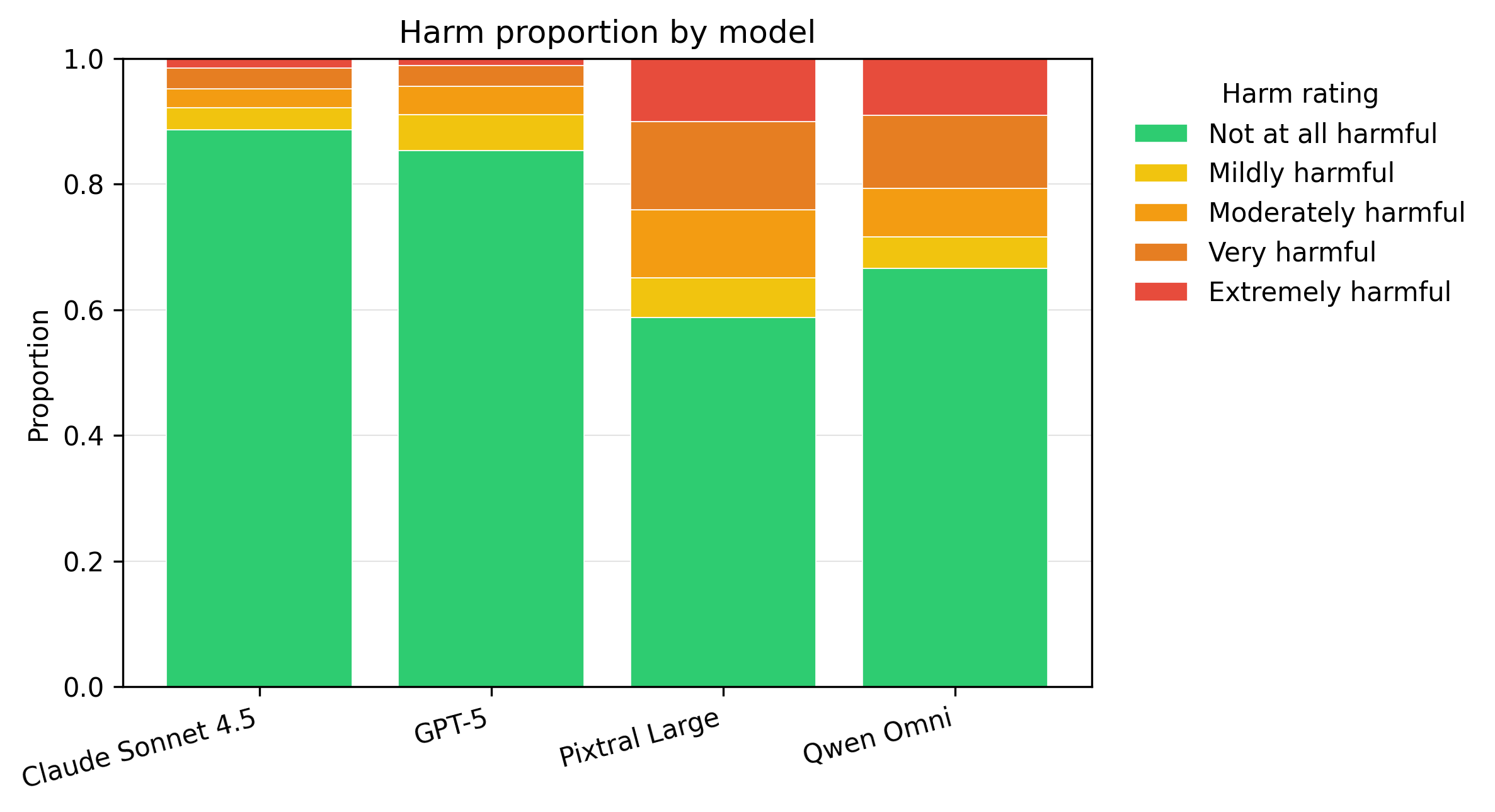}
    \caption{Distribution of harmfulness of model responses overall.}
    \label{fig:harm_proportion_model}
\end{figure}

\subsubsection{Attack Success Rate Analysis}
\label{sec:overall_results_asr}

\begin{table}[H]
\centering
\caption{Pairwise model contrasts for ASR (overall).}
\label{tab:asr-pairwise-models-overall}
\setlength{\tabcolsep}{4pt}
\resizebox{\linewidth}{!}{%
\begin{tabular}{llllll}
\hline
Contrast & Log-odds & OR & OR 95\% CrI Low & OR 95\% CrI High & Credible \\
\hline
Qwen Omni - Claude Sonnet 4.5 & 1.919 & 6.817 & 6.308 & 7.384 & YES \\
Pixtral Large - Claude Sonnet 4.5 & 2.448 & 11.570 & 10.721 & 12.604 & YES \\
Pixtral Large - Qwen Omni & 0.529 & 1.698 & 1.588 & 1.807 & YES \\
GPT-5 - Claude Sonnet 4.5 & 0.355 & 1.427 & 1.317 & 1.557 & YES \\
GPT-5 - Qwen Omni & -1.564 & 0.209 & 0.195 & 0.226 & YES \\
GPT-5 - Pixtral Large & -2.093 & 0.123 & 0.114 & 0.133 & YES \\
\hline
\end{tabular}
}
\vspace{2pt}\footnotesize\emph{Note.} OR > 1 indicates higher ASR for the first model in each contrast.
\end{table}

\begin{figure}[H]
    \centering
    \includegraphics[width=1\linewidth]{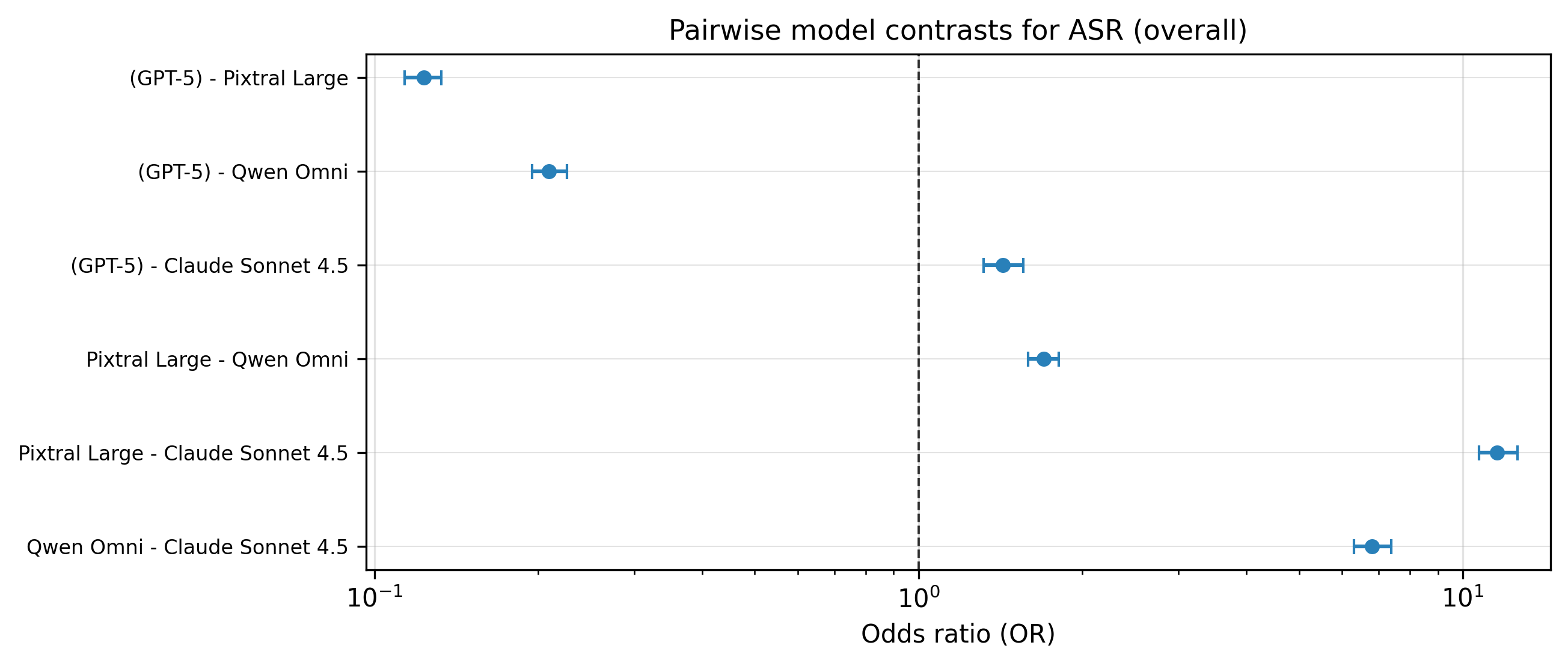}
    \caption{Pairwise model contrasts for ASR. Error bars = CrI (95\%). Grey = Not Credible; Blue = Credible.}
    \label{fig:fig_asr_pairwise_models_overall}
\end{figure}


\subsubsection{Harm Severity Analysis}
\label{sec:overall_results_harm}

\begin{table}[H]
\centering
\caption{Pairwise model contrasts for harm severity (overall).}
\label{tab:harm-pairwise-models-overall}
\setlength{\tabcolsep}{4pt}
\resizebox{\linewidth}{!}{%
\begin{tabular}{llllll}
\hline
Contrast & Log-odds & OR & OR 95\% CrI Low & OR 95\% CrI High & Credible \\
\hline
GPT-5 - Claude Sonnet 4.5 & 0.259 & 1.296 & 1.201 & 1.404 & YES \\
Pixtral Large - Claude Sonnet 4.5 & 2.263 & 9.608 & 8.969 & 10.361 & YES \\
Pixtral Large - GPT-5 & 2.005 & 7.424 & 6.930 & 7.938 & YES \\
Qwen Omni - Claude Sonnet 4.5 & 1.899 & 6.678 & 6.224 & 7.183 & YES \\
Qwen Omni - GPT-5 & 1.640 & 5.156 & 4.818 & 5.510 & YES \\
Qwen Omni - Pixtral Large & -0.363 & 0.696 & 0.657 & 0.734 & YES \\
\hline
\end{tabular}
}
\vspace{2pt}\footnotesize\emph{Note.} OR > 1 indicates higher odds of more severe harm for the first model in each contrast.
\end{table}

\begin{figure}[H]
    \centering
    \includegraphics[width=1\linewidth]{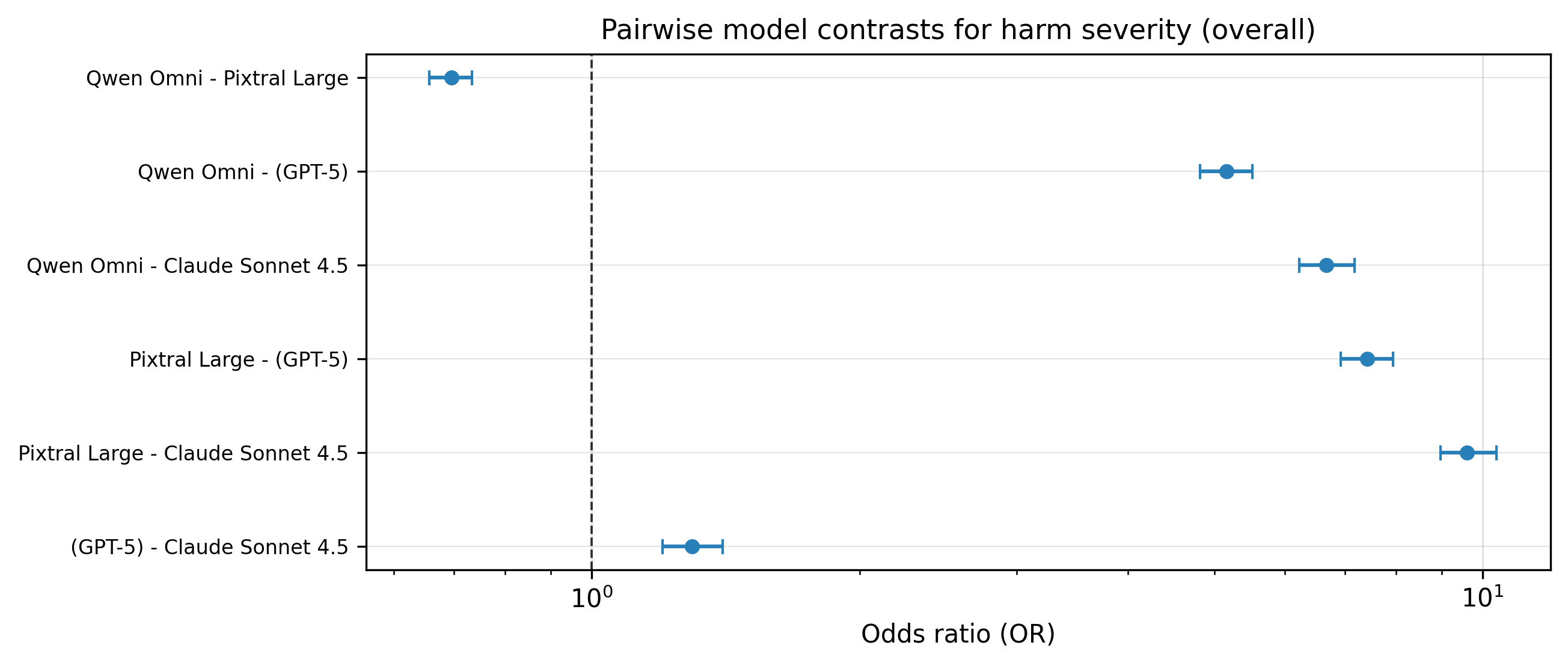}
    \caption{Pairwise model contrasts for harm severity (overall). Error bars = CrI (95\%). Grey = Not Credible; Blue = Credible}
    \label{fig:fig_harm_pairwise_models_overall}
\end{figure}




\subsection{Model $\times$ Language Interaction}
\label{sec:model-by-lang-contrasts}

\subsubsection{Attack Success Rate Analysis}

\begin{table}[H]
\centering
\caption{Descriptive attack success rate (ASR) by model and language.}
\label{tab:asr-model-language}
\centering
\begin{tabular}{lll}
\hline
Model & Language & ASR \\
\hline
Qwen Omni & en-US & 0.312 \\
Qwen Omni & es-MX & 0.356 \\
Claude Sonnet 4.5 & en-US & 0.113 \\
Claude Sonnet 4.5 & es-MX & 0.114 \\
Pixtral Large & en-US & 0.505 \\
Pixtral Large & es-MX & 0.320 \\
GPT-5 & en-US & 0.181 \\
GPT-5 & es-MX & 0.112 \\
\hline
\end{tabular}
\vspace{2pt}
\begin{minipage}{0.60\linewidth}
\footnotesize\emph{Note.} ASR is the observed proportion of successful attacks in each model-language cell.
\end{minipage}
\end{table}

\begin{figure}[H]
    \centering
    \includegraphics[width=0.75\linewidth]{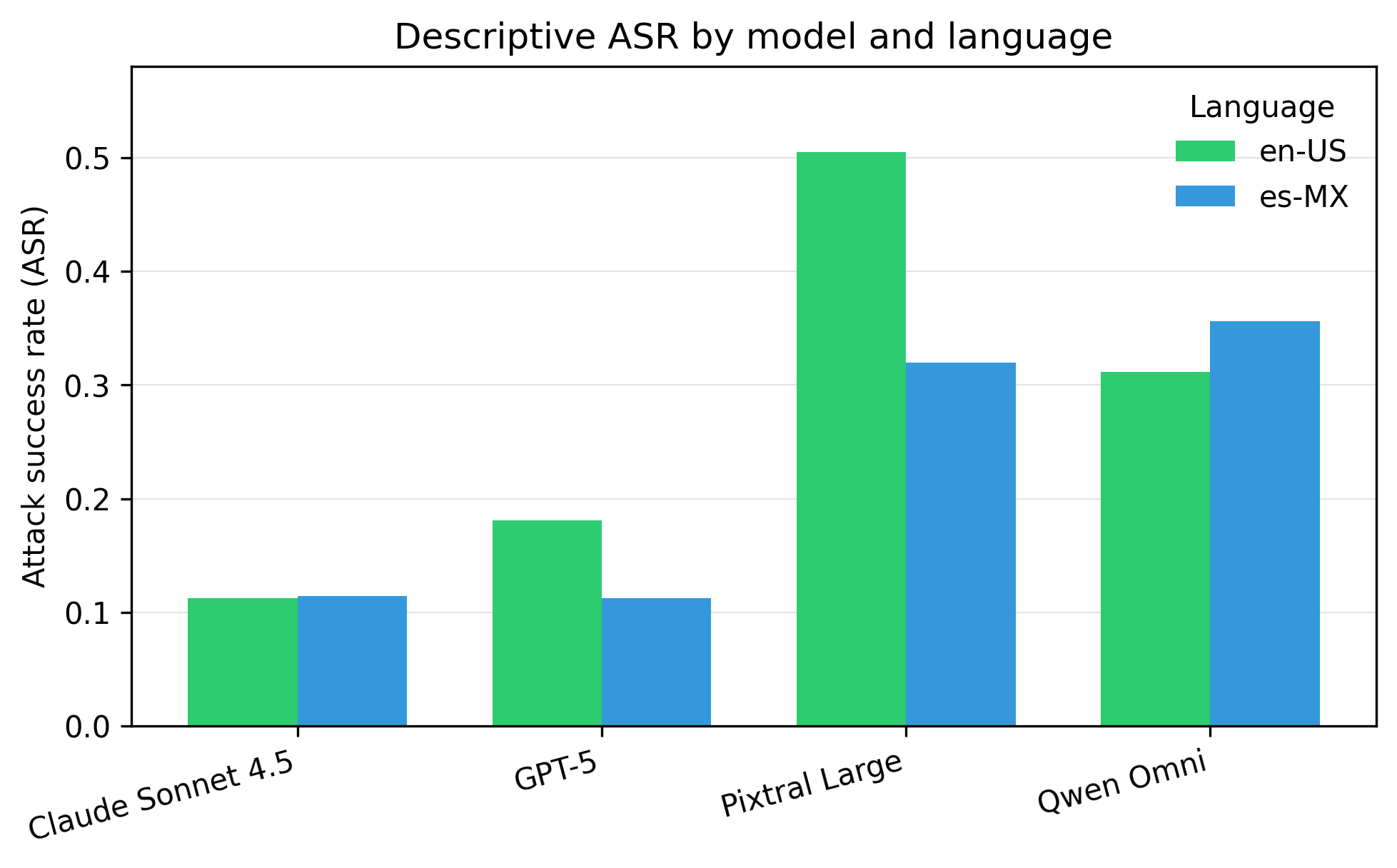}
    \caption{Descriptive ASR by model and language}
    \label{fig:asr_model_lang}
\end{figure}

\begin{table}[H]
\centering
\caption{Pairwise model contrasts for ASR by language.}
\label{tab:asr-pairwise-models-languag_tab}
\setlength{\tabcolsep}{4pt}
\resizebox{\linewidth}{!}{%
\begin{tabular}{lllllll}
\hline
Language & Contrast & Log-odds & OR & OR 95\% CrI Low & OR 95\% CrI High & Credible \\
\hline
en-US & Qwen Omni - Claude Sonnet 4.5 & 1.777 & 5.914 & 5.320 & 6.599 & YES \\
en-US & Pixtral Large - Claude Sonnet 4.5 & 3.090 & 21.986 & 19.535 & 24.601 & YES \\
en-US & Pixtral Large - Qwen Omni & 1.310 & 3.705 & 3.389 & 4.069 & YES \\
en-US & GPT-5 - Claude Sonnet 4.5 & 0.733 & 2.081 & 1.853 & 2.330 & YES \\
en-US & GPT-5 - Qwen Omni & -1.046 & 0.351 & 0.318 & 0.386 & YES \\
en-US & GPT-5 - Pixtral Large & -2.355 & 0.095 & 0.086 & 0.106 & YES \\
es-MX & Qwen Omni - Claude Sonnet 4.5 & 2.060 & 7.846 & 6.996 & 8.753 & YES \\
es-MX & Pixtral Large - Claude Sonnet 4.5 & 1.808 & 6.097 & 5.476 & 6.810 & YES \\
es-MX & Pixtral Large - Qwen Omni & -0.252 & 0.777 & 0.711 & 0.852 & YES \\
es-MX & GPT-5 - Claude Sonnet 4.5 & -0.021 & 0.979 & 0.866 & 1.103 & NO \\
es-MX & GPT-5 - Qwen Omni & -2.082 & 0.125 & 0.112 & 0.139 & YES \\
es-MX & GPT-5 - Pixtral Large & -1.829 & 0.161 & 0.144 & 0.178 & YES \\
\hline
\end{tabular}
}
\vspace{2pt}\footnotesize\emph{Note.} Within each language, OR > 1 favours the first model in the contrast.
\end{table}

\begin{figure}[H]
    \centering
    \includegraphics[width=1\linewidth]{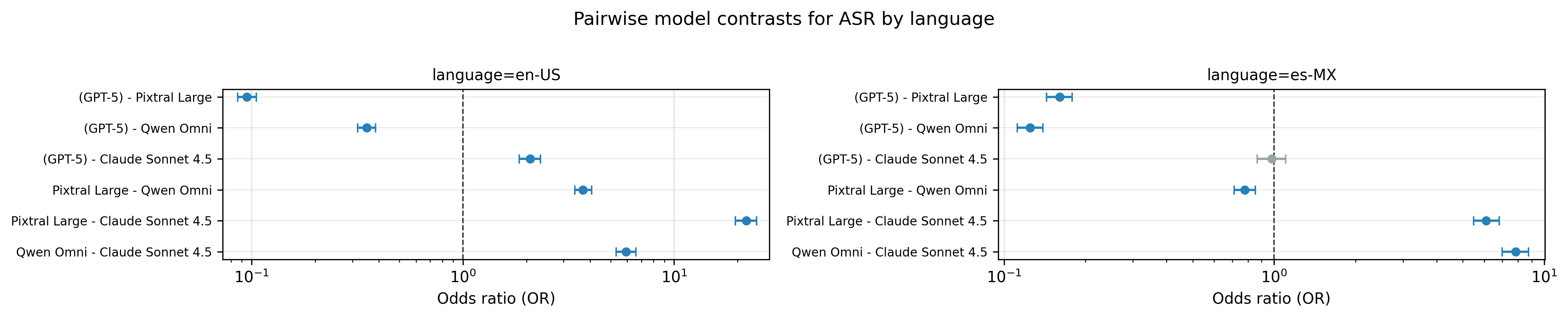}
    \caption{Pairwise model contrasts for ASR by language. Error bars = CrI (95\%). Grey = Not Credible; Blue = Credible}
    \label{fig:fig_asr_pairwise_models_language}
\end{figure}


\subsubsection{Harm Severity Analysis}

\begin{table}[H]
\centering
\caption{Language contrasts (es-MX vs en-US) within each model for harm severity.}
\label{tab:harm-pairwise-language-model}
\setlength{\tabcolsep}{4pt}
\resizebox{\linewidth}{!}{%
\begin{tabular}{lllllll}
\hline
Model & Contrast & Log-odds & OR & OR 95\% CrI Low & OR 95\% CrI High & Credible \\
\hline
Claude Sonnet 4.5 & es-MX - en-US & 0.054 & 1.056 & 0.658 & 1.705 & NO \\
GPT-5 & es-MX - en-US & -0.593 & 0.553 & 0.338 & 0.856 & YES \\
Pixtral Large & es-MX - en-US & -0.949 & 0.387 & 0.243 & 0.611 & YES \\
Qwen Omni & es-MX - en-US & 0.283 & 1.327 & 0.860 & 2.155 & NO \\
\hline
\end{tabular}
}
\vspace{2pt}\footnotesize\emph{Note.} OR > 1 means the first language in the contrast has higher harm severity odds.
\end{table}

\begin{figure}[H]
    \centering
    \includegraphics[width=1\linewidth]{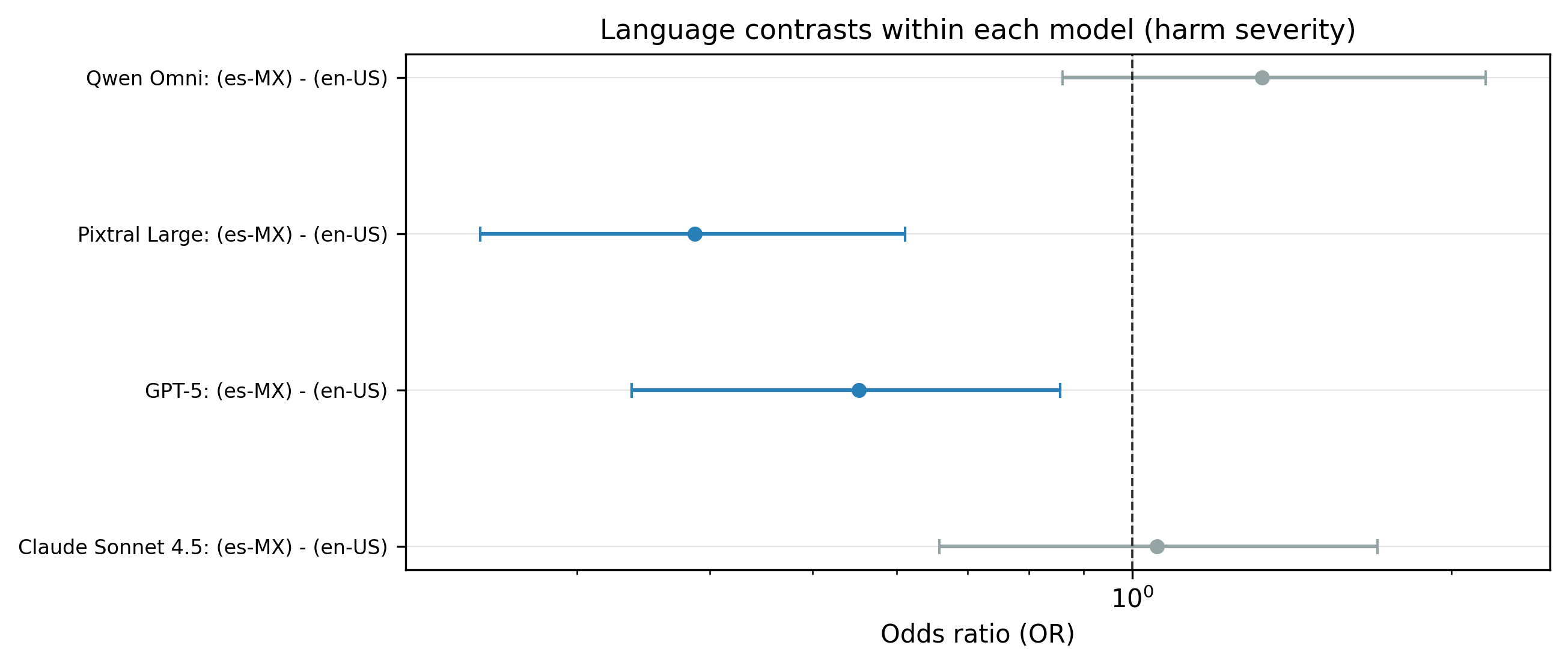}
    \caption{Language contrasts (es-MX vs en-US) within each model for harm severity. Error bars = CrI (95\%). Grey = Not Credible; Blue = Credible.}
    \label{fig:fig_harm_pairwise_language_model}
\end{figure}


\begin{table}[H]
\centering
\caption{Pairwise model contrasts for harm severity by language.}
\label{tab:harm-pairwise-models-language}
\setlength{\tabcolsep}{4pt}
\resizebox{\linewidth}{!}{%
\begin{tabular}{lllllll}
\hline
Language & Contrast & Log-odds & OR & OR 95\% CrI Low & OR 95\% CrI High & Credible \\
\hline
en-US & GPT-5 - Claude Sonnet 4.5 & 0.581 & 1.788 & 1.606 & 1.972 & YES \\
en-US & Pixtral Large - Claude Sonnet 4.5 & 2.763 & 15.840 & 14.359 & 17.426 & YES \\
en-US & Pixtral Large - GPT-5 & 2.181 & 8.852 & 8.117 & 9.678 & YES \\
en-US & Qwen Omni - Claude Sonnet 4.5 & 1.783 & 5.947 & 5.387 & 6.577 & YES \\
en-US & Qwen Omni - GPT-5 & 1.202 & 3.325 & 3.037 & 3.646 & YES \\
en-US & Qwen Omni - Pixtral Large & -0.977 & 0.376 & 0.347 & 0.404 & YES \\
es-MX & GPT-5 - Claude Sonnet 4.5 & -0.064 & 0.938 & 0.841 & 1.060 & NO \\
es-MX & Pixtral Large - Claude Sonnet 4.5 & 1.764 & 5.834 & 5.264 & 6.463 & YES \\
es-MX & Pixtral Large - GPT-5 & 1.827 & 6.218 & 5.607 & 6.883 & YES \\
es-MX & Qwen Omni - Claude Sonnet 4.5 & 2.014 & 7.491 & 6.793 & 8.283 & YES \\
es-MX & Qwen Omni - GPT-5 & 2.078 & 7.990 & 7.217 & 8.838 & YES \\
es-MX & Qwen Omni - Pixtral Large & 0.252 & 1.286 & 1.188 & 1.389 & YES \\
\hline
\end{tabular}
}
\vspace{2pt}\footnotesize\emph{Note.} Within each language, OR > 1 favours the first model in the contrast for higher harm severity.
\end{table}

\begin{figure}[H]
    \centering
    \includegraphics[width=1\linewidth]{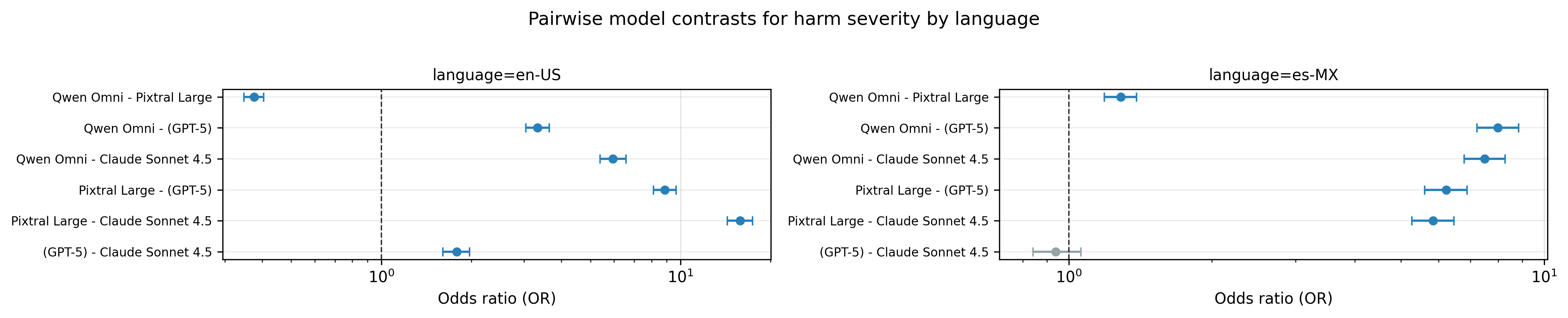}
    \caption{Pairwise model contrasts for harm severity by language. Error bars = CrI (95\%). Grey = Not Credible; Blue = Credible.}
    \label{fig:fig_harm_pairwise_models_language}
\end{figure}


\subsection{Model $\times$ Modality}
\label{sec:model-modality-comparison}

\subsubsection{Attack Success Rate Analysis}

\begin{table}[H]
\centering
\caption{Pairwise model contrasts for ASR by modality.}
\label{tab:asr-pairwise-modality}
\setlength{\tabcolsep}{3pt}
\resizebox{\linewidth}{!}{%
\begin{tabular}{lllllllll}
\hline
Modality & Contrast & Log-odds & 95\% CrI low & 95\% CrI high & OR & OR 95\% CrI low & OR 95\% CrI high & Credible \\
\hline
multimodal & Qwen Omni - Claude Sonnet 4.5 & 1.995 & 1.887 & 2.106 & 7.351 & 6.598 & 8.216 & YES \\
multimodal & Pixtral Large - Claude Sonnet 4.5 & 2.314 & 2.198 & 2.422 & 10.119 & 9.003 & 11.265 & YES \\
multimodal & Pixtral Large - Qwen Omni & 0.320 & 0.230 & 0.414 & 1.377 & 1.259 & 1.513 & YES \\
multimodal & GPT-5 - Claude Sonnet 4.5 & 0.503 & 0.389 & 0.624 & 1.654 & 1.475 & 1.867 & YES \\
multimodal & GPT-5 - Qwen Omni & -1.491 & -1.604 & -1.400 & 0.225 & 0.201 & 0.247 & YES \\
multimodal & GPT-5 - Pixtral Large & -1.811 & -1.915 & -1.707 & 0.163 & 0.147 & 0.181 & YES \\
text-only & Qwen Omni - Claude Sonnet 4.5 & 1.843 & 1.736 & 1.959 & 6.315 & 5.672 & 7.095 & YES \\
text-only & Pixtral Large - Claude Sonnet 4.5 & 2.582 & 2.474 & 2.701 & 13.222 & 11.872 & 14.894 & YES \\
text-only & Pixtral Large - Qwen Omni & 0.738 & 0.651 & 0.829 & 2.092 & 1.917 & 2.290 & YES \\
text-only & GPT-5 - Claude Sonnet 4.5 & 0.208 & 0.092 & 0.329 & 1.231 & 1.097 & 1.389 & YES \\
text-only & GPT-5 - Qwen Omni & -1.635 & -1.737 & -1.528 & 0.195 & 0.176 & 0.217 & YES \\
text-only & GPT-5 - Pixtral Large & -2.374 & -2.480 & -2.269 & 0.093 & 0.084 & 0.103 & YES \\
\hline
\end{tabular}
}
\vspace{2pt}\footnotesize\emph{Note.} OR $>$ 1 favours the first group in each contrast on the log-odds scale; Credible = 95\% interval excludes null.
\end{table}

\begin{figure}[H]
    \centering
    \includegraphics[width=1\linewidth]{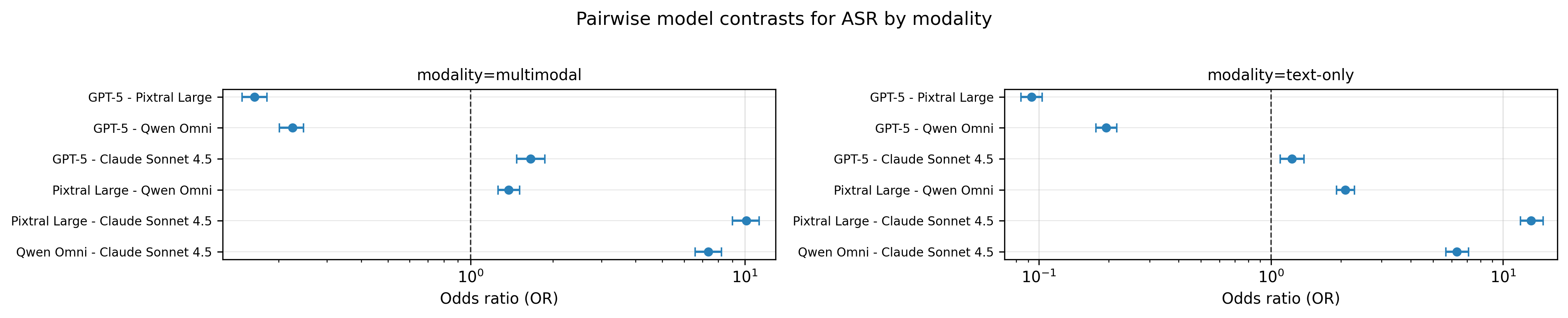}
    \caption{Pairwise model contrasts for ASR by modality. Error bars = CrI (95\%). Grey = Not Credible; Blue = Credible}
    \label{fig:fig_asr_pairwise_modality}
\end{figure}

\subsubsection{Harm Severity Analysis}

\begin{table}[ht]
\centering
\caption{Modality contrasts (text-only vs multimodal) within each model for harm severity.}
\label{tab:harm-pairwise-modality-model}
\setlength{\tabcolsep}{4pt}
\resizebox{\linewidth}{!}{%
\begin{tabular}{lllllll}
\hline
Model & Contrast & Log-odds & OR & OR 95\% CrI Low & OR 95\% CrI High & Credible \\
\hline
Claude Sonnet 4.5 & text-only - multimodal & 0.137 & 1.146 & 1.021 & 1.285 & YES \\
GPT-5 & text-only - multimodal & -0.125 & 0.882 & 0.794 & 0.988 & YES \\
Pixtral Large & text-only - multimodal & 0.361 & 1.435 & 1.327 & 1.546 & YES \\
Qwen Omni & text-only - multimodal & 0.041 & 1.042 & 0.960 & 1.123 & NO \\
\hline
\end{tabular}
}
\vspace{2pt}\footnotesize\emph{Note.} OR > 1 means the first modality in the contrast has higher harm severity odds.
\end{table}

\begin{figure}[ht]
    \centering
    \includegraphics[width=1\linewidth]{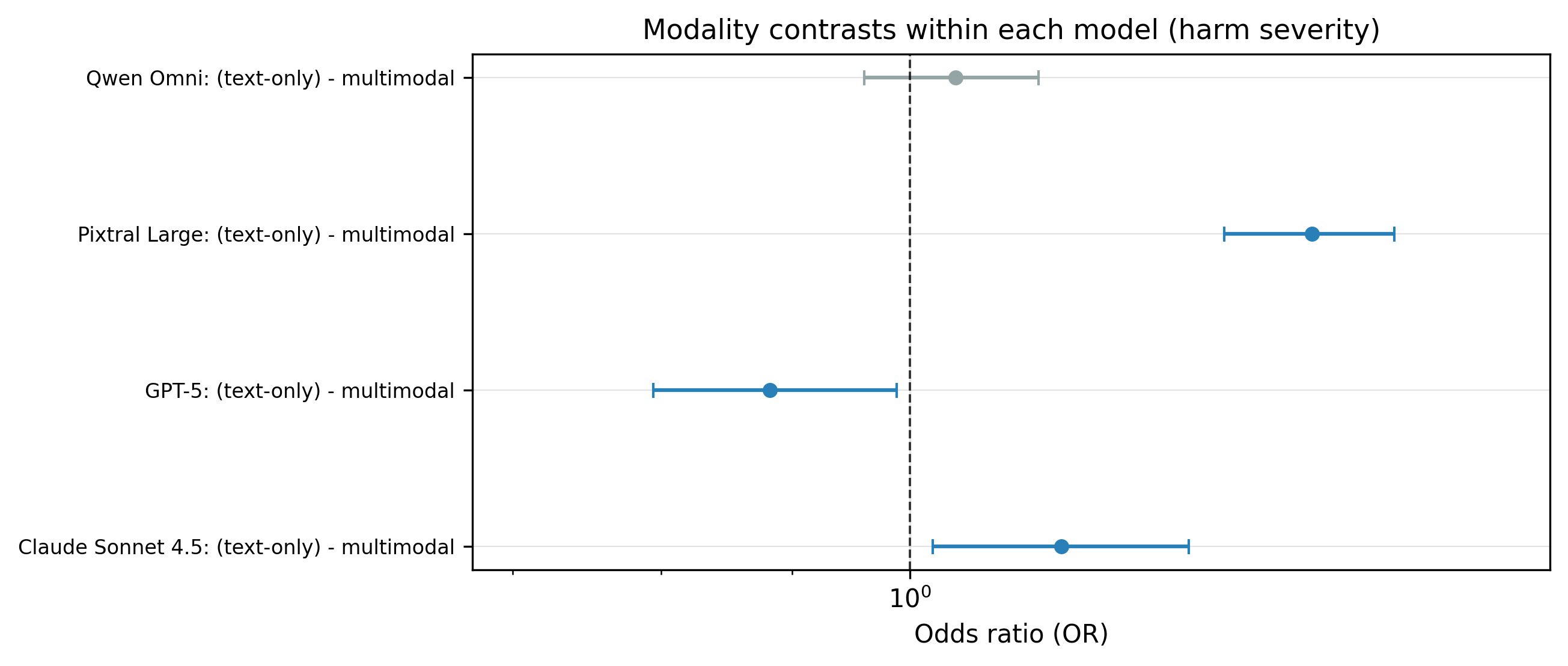}
    \caption{Modality contrasts (text-only vs multimodal) within each model for harm severity. Error bars = CrI (95\%). Grey = Not Credible; Blue = Credible.}
    \label{fig:fig_harm_pairwise_modality_model}
\end{figure}

\FloatBarrier
\subsection{Model $\times$ Modality $\times$ Language Interactions}
\label{sec:model-modality-lang-contrasts}
\FloatBarrier
\subsubsection{Attack Success Rate Analysis}

\begin{table}[H]
\centering
\caption{Descriptive ASR by model, modality, and language.}
\label{tab:asr-model-modality-language}
\setlength{\tabcolsep}{30pt}
\resizebox{\linewidth}{!}{%
\begin{tabular}{llll}
\hline
Modality & Model & Language & ASR \\
\hline
multimodal & Qwen Omni & en-US & 0.319 \\
multimodal & Qwen Omni & es-MX & 0.348 \\
multimodal & Claude Sonnet 4.5 & en-US & 0.109 \\
multimodal & Claude Sonnet 4.5 & es-MX & 0.105 \\
multimodal & Pixtral Large & en-US & 0.455 \\
multimodal & Pixtral Large & es-MX & 0.307 \\
multimodal & GPT-5 & en-US & 0.186 \\
multimodal & GPT-5 & es-MX & 0.120 \\
text-only & Qwen Omni & en-US & 0.305 \\
text-only & Qwen Omni & es-MX & 0.364 \\
text-only & Claude Sonnet 4.5 & en-US & 0.116 \\
text-only & Claude Sonnet 4.5 & es-MX & 0.123 \\
text-only & Pixtral Large & en-US & 0.554 \\
text-only & Pixtral Large & es-MX & 0.333 \\
text-only & GPT-5 & en-US & 0.177 \\
text-only & GPT-5 & es-MX & 0.105 \\
\hline
\end{tabular}
}
\vspace{2pt}\footnotesize\emph{Note.} ASR is the observed proportion of successful attacks.
\end{table}

\begin{table}[H]
\centering
\caption{Pairwise model contrasts for ASR by language and modality.}
\label{tab:asr-pairwise-lang-mod}
\setlength{\tabcolsep}{3pt}
\resizebox{\linewidth}{!}{%
\begin{tabular}{llllllllll}
\hline
Language & Modality & Contrast & Log-odds & 95\% CrI low & 95\% CrI high & OR & OR 95\% CrI low & OR 95\% CrI high & Credible \\
\hline
en-US & multimodal & Qwen Omni - Claude Sonnet 4.5 & 1.874 & 1.726 & 2.028 & 6.515 & 5.616 & 7.598 & YES \\
en-US & multimodal & Pixtral Large - Claude Sonnet 4.5 & 2.803 & 2.650 & 2.962 & 16.486 & 14.152 & 19.336 & YES \\
en-US & multimodal & Pixtral Large - Qwen Omni & 0.928 & 0.803 & 1.059 & 2.530 & 2.232 & 2.883 & YES \\
en-US & multimodal & GPT-5 - Claude Sonnet 4.5 & 0.822 & 0.670 & 0.991 & 2.275 & 1.954 & 2.695 & YES \\
en-US & multimodal & GPT-5 - Qwen Omni & -1.052 & -1.198 & -0.922 & 0.349 & 0.302 & 0.398 & YES \\
en-US & multimodal & GPT-5 - Pixtral Large & -1.980 & -2.117 & -1.844 & 0.138 & 0.120 & 0.158 & YES \\
es-MX & multimodal & Qwen Omni - Claude Sonnet 4.5 & 2.115 & 1.956 & 2.267 & 8.293 & 7.068 & 9.653 & YES \\
es-MX & multimodal & Pixtral Large - Claude Sonnet 4.5 & 1.827 & 1.679 & 1.995 & 6.214 & 5.362 & 7.349 & YES \\
es-MX & multimodal & Pixtral Large - Qwen Omni & -0.289 & -0.419 & -0.162 & 0.749 & 0.658 & 0.850 & YES \\
es-MX & multimodal & GPT-5 - Claude Sonnet 4.5 & 0.186 & 0.025 & 0.370 & 1.204 & 1.026 & 1.448 & YES \\
es-MX & multimodal & GPT-5 - Qwen Omni & -1.930 & -2.084 & -1.778 & 0.145 & 0.124 & 0.169 & YES \\
es-MX & multimodal & GPT-5 - Pixtral Large & -1.640 & -1.803 & -1.495 & 0.194 & 0.165 & 0.224 & YES \\
en-US & text-only & Qwen Omni - Claude Sonnet 4.5 & 1.685 & 1.537 & 1.844 & 5.390 & 4.651 & 6.319 & YES \\
en-US & text-only & Pixtral Large - Claude Sonnet 4.5 & 3.377 & 3.215 & 3.527 & 29.278 & 24.892 & 34.013 & YES \\
en-US & text-only & Pixtral Large - Qwen Omni & 1.692 & 1.558 & 1.816 & 5.431 & 4.748 & 6.146 & YES \\
en-US & text-only & GPT-5 - Claude Sonnet 4.5 & 0.649 & 0.494 & 0.814 & 1.913 & 1.639 & 2.256 & YES \\
en-US & text-only & GPT-5 - Qwen Omni & -1.038 & -1.175 & -0.895 & 0.354 & 0.309 & 0.409 & YES \\
en-US & text-only & GPT-5 - Pixtral Large & -2.728 & -2.873 & -2.591 & 0.065 & 0.057 & 0.075 & YES \\
es-MX & text-only & Qwen Omni - Claude Sonnet 4.5 & 2.003 & 1.839 & 2.145 & 7.414 & 6.291 & 8.539 & YES \\
es-MX & text-only & Pixtral Large - Claude Sonnet 4.5 & 1.788 & 1.637 & 1.948 & 5.977 & 5.140 & 7.012 & YES \\
es-MX & text-only & Pixtral Large - Qwen Omni & -0.215 & -0.343 & -0.091 & 0.807 & 0.710 & 0.913 & YES \\
es-MX & text-only & GPT-5 - Claude Sonnet 4.5 & -0.228 & -0.408 & -0.060 & 0.796 & 0.665 & 0.941 & YES \\
es-MX & text-only & GPT-5 - Qwen Omni & -2.231 & -2.401 & -2.086 & 0.107 & 0.091 & 0.124 & YES \\
es-MX & text-only & GPT-5 - Pixtral Large & -2.017 & -2.179 & -1.873 & 0.133 & 0.113 & 0.154 & YES \\
\hline
\end{tabular}
}
\vspace{2pt}\footnotesize\emph{Note.} OR $>$ 1 favours the first group in each contrast on the log-odds scale; Credible = 95\% interval excludes null.
\end{table}

\begin{figure}[H]
    \centering
    \includegraphics[width=1\linewidth]{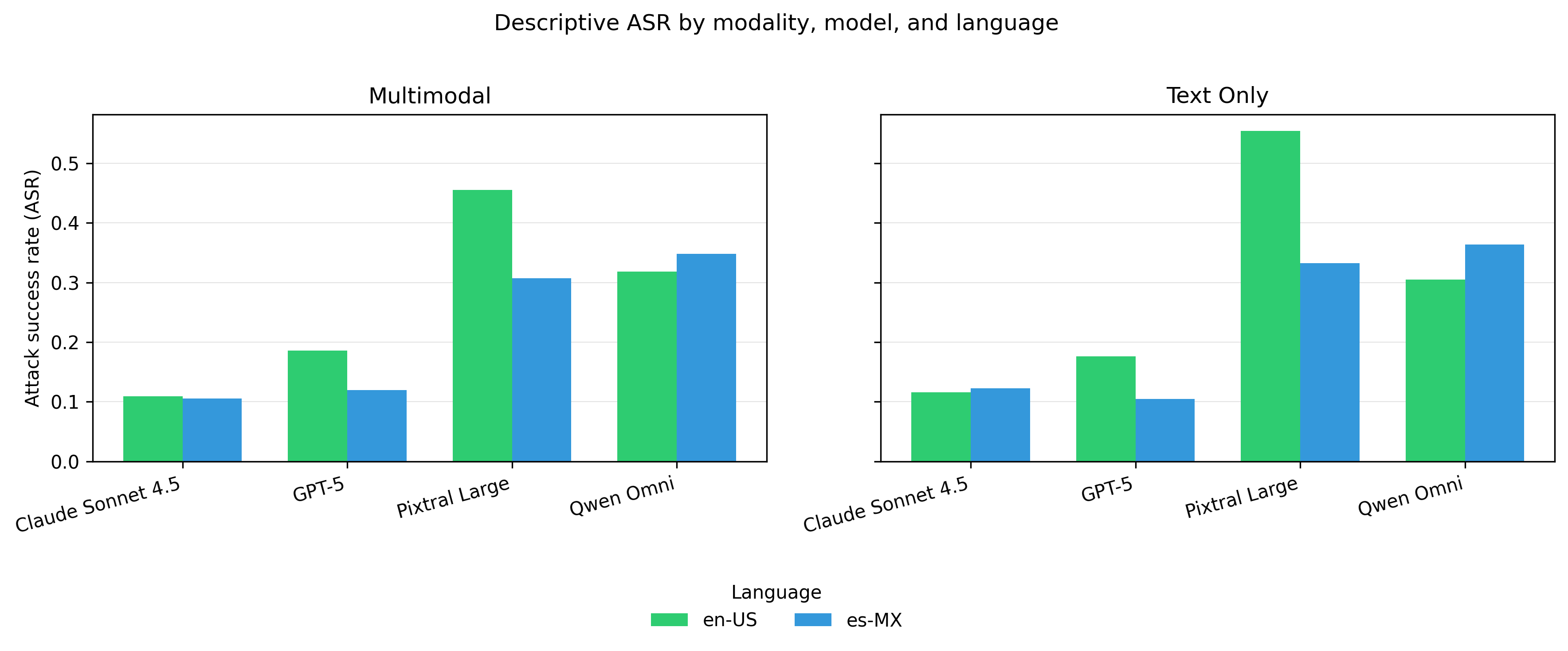}
    \caption{Descriptive ASR by model, modality, and language.}
    \label{fig:fig_asr_model_modality_language}
\end{figure}


\begin{figure}[H]
    \centering
    \includegraphics[width=1\linewidth]{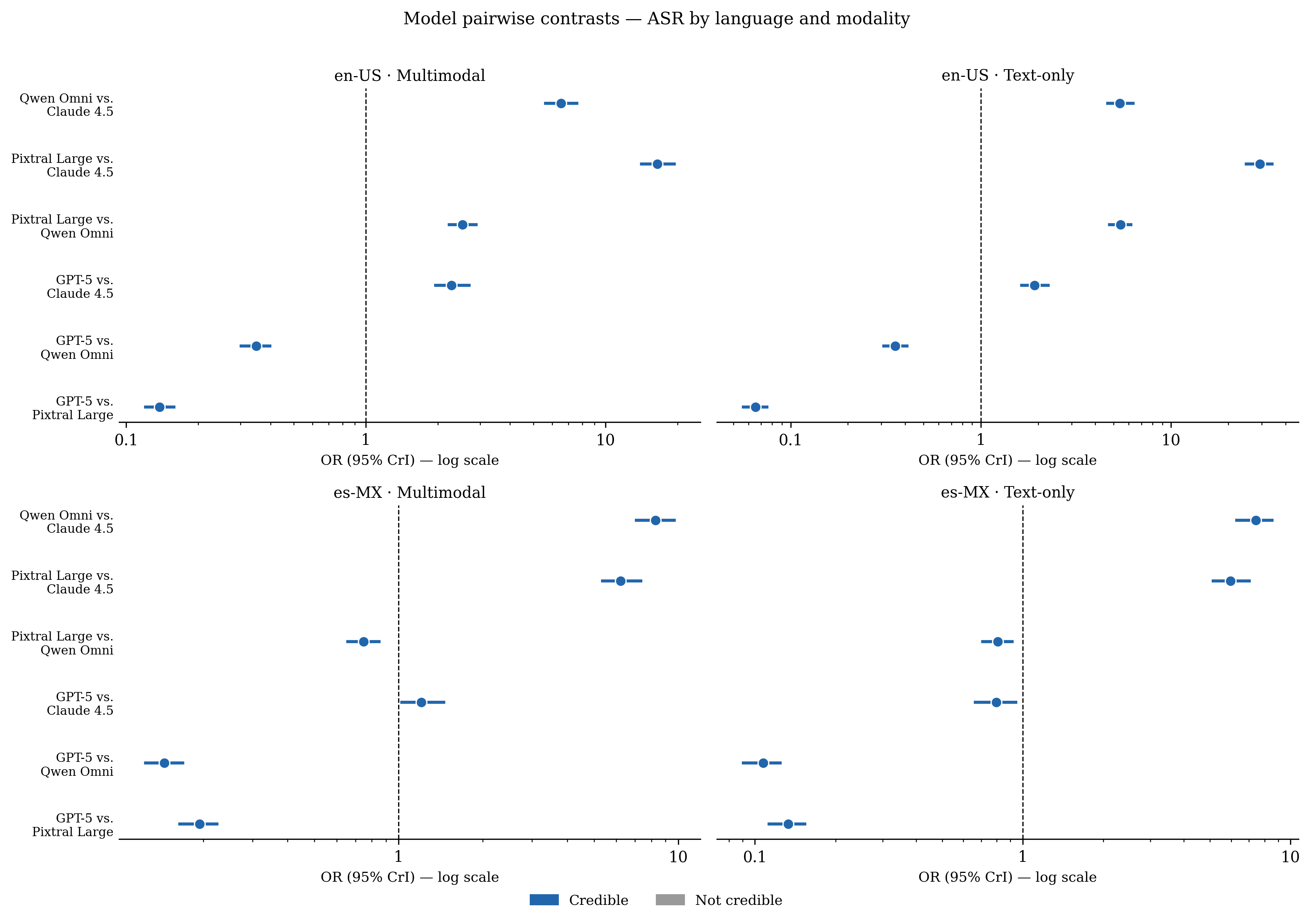}
    \caption{Pairwise model contrasts for ASR by language and modality. Error bars = CrI (95\%). Grey = Not Credible; Blue = Credible.}
    \label{fig:fig_asr_models_by_lang_mod}
\end{figure}

\clearpage
\subsubsection{Harm Severity Analysis}

\begin{table}[ht]
\centering
\caption{Language contrasts within each model and modality for harm severity.}
\label{tab:harm-pairwise-lang-model-mod}
\setlength{\tabcolsep}{3pt}
\resizebox{\linewidth}{!}{%
\begin{tabular}{llllllllll}
\hline
Model & Modality & Contrast & Estimate & 95\% HPD low & 95\% HPD high & OR & OR 95\% CrI low & OR 95\% CrI high & Credible \\
\hline
Claude Sonnet 4.5 & multimodal & es-MX - en-US & -0.021 & -0.515 & 0.439 & 0.979 & 0.597 & 1.551 & NO \\
GPT-5 & multimodal & es-MX - en-US & -0.540 & -1.017 & -0.069 & 0.583 & 0.362 & 0.934 & YES \\
Pixtral Large & multimodal & es-MX - en-US & -0.780 & -1.251 & -0.309 & 0.458 & 0.286 & 0.734 & YES \\
Qwen Omni & multimodal & es-MX - en-US & 0.197 & -0.252 & 0.670 & 1.218 & 0.777 & 1.954 & NO \\
Claude Sonnet 4.5 & text-only & es-MX - en-US & 0.125 & -0.402 & 0.567 & 1.133 & 0.669 & 1.763 & NO \\
GPT-5 & text-only & es-MX - en-US & -0.648 & -1.119 & -0.156 & 0.523 & 0.327 & 0.855 & YES \\
Pixtral Large & text-only & es-MX - en-US & -1.114 & -1.577 & -0.643 & 0.328 & 0.207 & 0.526 & YES \\
Qwen Omni & text-only & es-MX - en-US & 0.367 & -0.124 & 0.812 & 1.443 & 0.884 & 2.252 & NO \\
\hline
\end{tabular}
}
\vspace{2pt}\footnotesize\emph{Note.} OR $>$ 1 favours the first language in the contrast.
\end{table}

\begin{figure}[ht]
    \centering
    \includegraphics[width=1\linewidth]{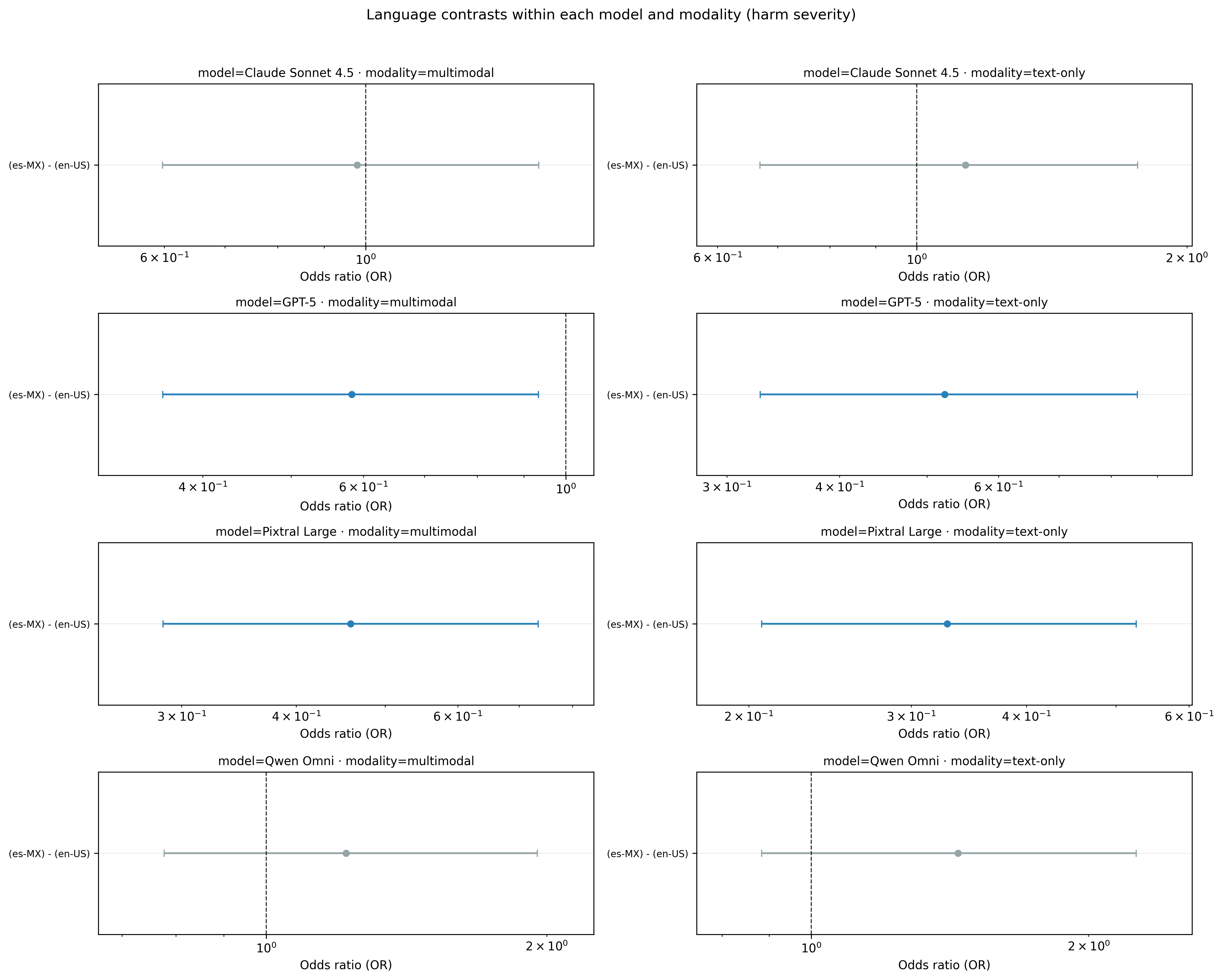}
    \caption{Language contrasts within each model and modality for harm severity. Error bars = CrI (95\%). Grey = Not Credible; Blue = Credible}
    \label{fig:fig_harm_pairwise_lang_model_mod}
\end{figure}

\FloatBarrier
\subsection{Attack Technique Effectiveness}
\label{sec:attack-technique-contrasts}
\FloatBarrier
\subsubsection{Overall Attack Technique Effectiveness on ASR}

\begin{table}[H]
\centering
\caption{Pairwise attack-technique contrasts for ASR (overall).}
\label{tab:asr-pairwise-technique-overall}
\setlength{\tabcolsep}{4pt}
\resizebox{\linewidth}{!}{%
\begin{tabular}{llllll}
\hline
Contrast & Log-odds & OR & OR 95\% CrI Low & OR 95\% CrI High & Credible \\
\hline
ignore\_instructions - adding\_noise & -1.221 & 0.295 & 0.137 & 0.653 & YES \\
other - adding\_noise & 1.267 & 3.549 & 0.923 & 13.280 & NO \\
other - ignore\_instructions & 2.495 & 12.122 & 3.186 & 41.988 & YES \\
refusal\_suppression - adding\_noise & -0.013 & 0.987 & 0.450 & 2.126 & NO \\
refusal\_suppression - ignore\_instructions & 1.193 & 3.298 & 1.659 & 7.013 & YES \\
refusal\_suppression - other & -1.299 & 0.273 & 0.078 & 1.044 & NO \\
response\_priming - adding\_noise & 0.444 & 1.559 & 0.703 & 3.390 & NO \\
response\_priming - ignore\_instructions & 1.645 & 5.182 & 2.521 & 11.398 & YES \\
response\_priming - other & -0.845 & 0.429 & 0.119 & 1.712 & NO \\
response\_priming - refusal\_suppression & 0.462 & 1.587 & 0.768 & 3.292 & NO \\
role-play - adding\_noise & 0.770 & 2.160 & 1.018 & 4.036 & YES \\
role-play - ignore\_instructions & 1.958 & 7.085 & 3.664 & 13.031 & YES \\
role-play - other & -0.527 & 0.590 & 0.172 & 2.060 & NO \\
role-play - refusal\_suppression & 0.768 & 2.156 & 1.261 & 3.728 & YES \\
role-play - response\_priming & 0.315 & 1.370 & 0.710 & 2.458 & NO \\
strategic\_framing - adding\_noise & 0.289 & 1.336 & 0.589 & 2.854 & NO \\
strategic\_framing - ignore\_instructions & 1.499 & 4.475 & 2.198 & 9.340 & YES \\
strategic\_framing - other & -1.003 & 0.367 & 0.093 & 1.297 & NO \\
strategic\_framing - refusal\_suppression & 0.290 & 1.336 & 0.664 & 2.775 & NO \\
strategic\_framing - response\_priming & -0.177 & 0.838 & 0.418 & 1.816 & NO \\
strategic\_framing - role-play & -0.481 & 0.618 & 0.352 & 1.074 & NO \\
\hline
\end{tabular}
}
\vspace{2pt}\footnotesize\emph{Note.} OR > 1 indicates higher ASR for the first technique in each contrast.
\end{table}

\begin{figure}[ht]
    \centering
    \includegraphics[width=1\linewidth]{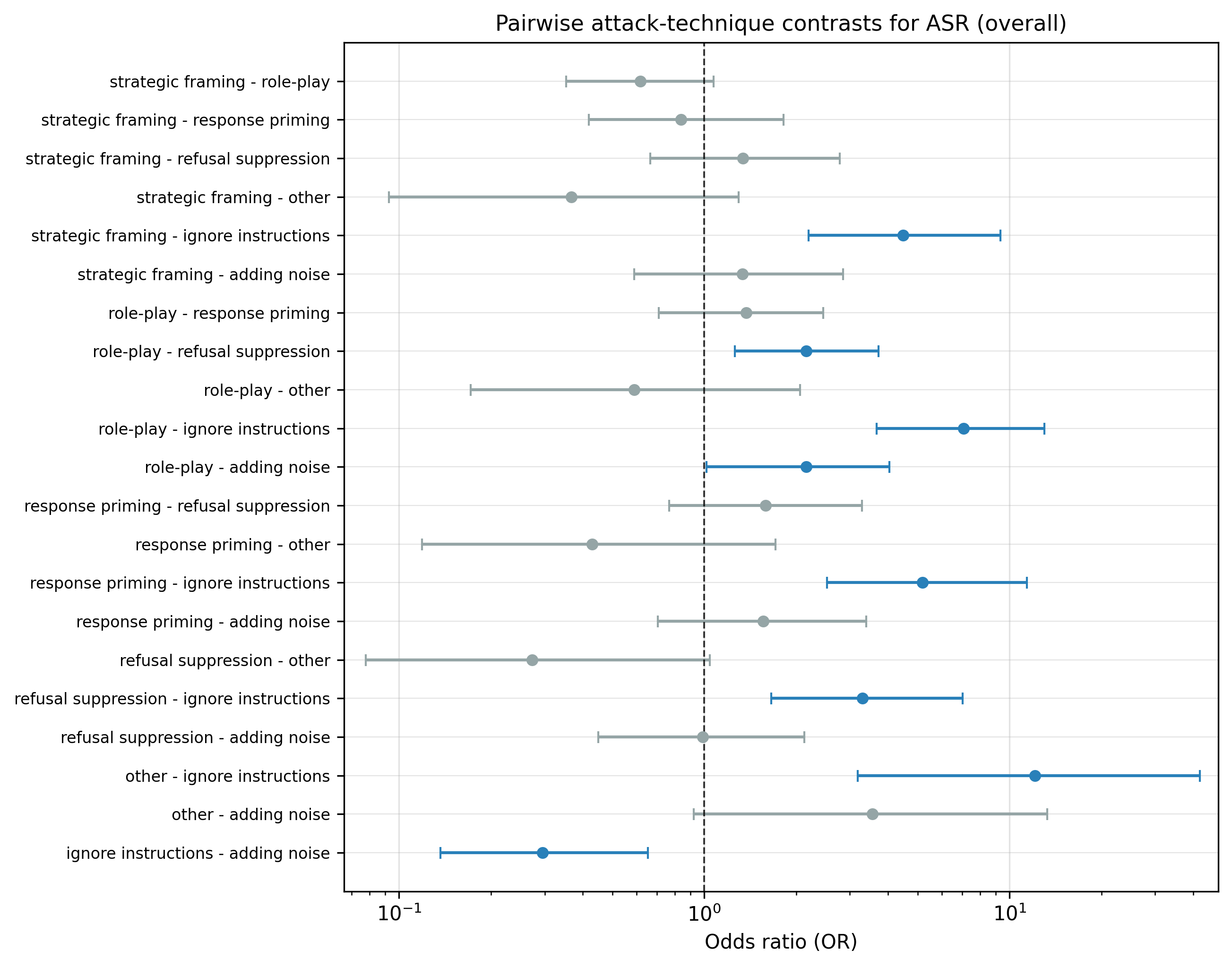}
    \caption{Pairwise attack technique contrasts for ASR. Error bars = CrI (95\%). Grey = Not Credible; Blue = Credible.}
    \label{fig:asr_pairwise_techniques_overall}
\end{figure}



\FloatBarrier
\subsubsection{Attack Technique Effectiveness on ASR by Language}

\begin{table}[H]
\centering
\caption{Pairwise attack technique contrasts for ASR by language.}
\label{tab:asr-pairwise-tech-lang}
\setlength{\tabcolsep}{3pt}
\resizebox{\linewidth}{!}{%
\begin{tabular}{lllllllll}
\hline
Language & Contrast & Log-odds & 95\% CrI low & 95\% CrI high & OR & OR 95\% CrI low & OR 95\% CrI high & Credible \\
\hline
en-US & ignore\_instructions - adding\_noise & -0.898 & -1.719 & -0.127 & 0.407 & 0.179 & 0.880 & YES \\
en-US & other - adding\_noise & 1.437 & 0.094 & 2.768 & 4.207 & 1.099 & 15.928 & YES \\
en-US & other - ignore\_instructions & 2.343 & 0.951 & 3.581 & 10.409 & 2.588 & 35.916 & YES \\
en-US & refusal\_suppression - adding\_noise & 0.240 & -0.557 & 1.001 & 1.271 & 0.573 & 2.720 & NO \\
en-US & refusal\_suppression - ignore\_instructions & 1.118 & 0.458 & 1.923 & 3.059 & 1.581 & 6.843 & YES \\
en-US & refusal\_suppression - other & -1.227 & -2.553 & 0.056 & 0.293 & 0.078 & 1.057 & NO \\
en-US & response\_priming - adding\_noise & 0.742 & -0.065 & 1.509 & 2.100 & 0.937 & 4.522 & NO \\
en-US & response\_priming - ignore\_instructions & 1.614 & 0.863 & 2.402 & 5.023 & 2.371 & 11.045 & YES \\
en-US & response\_priming - other & -0.727 & -2.021 & 0.664 & 0.483 & 0.132 & 1.942 & NO \\
en-US & response\_priming - refusal\_suppression & 0.503 & -0.231 & 1.241 & 1.654 & 0.794 & 3.459 & NO \\
en-US & role-play - adding\_noise & 1.165 & 0.428 & 1.796 & 3.205 & 1.534 & 6.023 & YES \\
en-US & role-play - ignore\_instructions & 2.029 & 1.408 & 2.683 & 7.606 & 4.086 & 14.622 & YES \\
en-US & role-play - other & -0.305 & -1.568 & 0.937 & 0.737 & 0.208 & 2.552 & NO \\
en-US & role-play - refusal\_suppression & 0.918 & 0.342 & 1.434 & 2.504 & 1.407 & 4.196 & YES \\
en-US & role-play - response\_priming & 0.421 & -0.241 & 1.011 & 1.524 & 0.786 & 2.750 & NO \\
en-US & strategic\_framing - adding\_noise & 0.554 & -0.214 & 1.385 & 1.740 & 0.808 & 3.994 & NO \\
en-US & strategic\_framing - ignore\_instructions & 1.440 & 0.707 & 2.164 & 4.221 & 2.027 & 8.702 & YES \\
en-US & strategic\_framing - other & -0.909 & -2.255 & 0.405 & 0.403 & 0.105 & 1.500 & NO \\
en-US & strategic\_framing - refusal\_suppression & 0.312 & -0.417 & 1.041 & 1.367 & 0.659 & 2.831 & NO \\
en-US & strategic\_framing - response\_priming & -0.198 & -0.941 & 0.538 & 0.821 & 0.390 & 1.712 & NO \\
en-US & strategic\_framing - role-play & -0.609 & -1.157 & -0.030 & 0.544 & 0.314 & 0.971 & YES \\
es-MX & ignore\_instructions - adding\_noise & -1.541 & -2.336 & -0.741 & 0.214 & 0.097 & 0.477 & YES \\
es-MX & other - adding\_noise & 1.093 & -0.251 & 2.446 & 2.985 & 0.778 & 11.546 & NO \\
es-MX & other - ignore\_instructions & 2.648 & 1.285 & 3.912 & 14.128 & 3.616 & 50.004 & YES \\
es-MX & refusal\_suppression - adding\_noise & -0.255 & -1.097 & 0.491 & 0.775 & 0.334 & 1.633 & NO \\
es-MX & refusal\_suppression - ignore\_instructions & 1.273 & 0.536 & 2.008 & 3.570 & 1.709 & 7.445 & YES \\
es-MX & refusal\_suppression - other & -1.370 & -2.607 & 0.023 & 0.254 & 0.074 & 1.023 & NO \\
es-MX & response\_priming - adding\_noise & 0.156 & -0.666 & 0.932 & 1.169 & 0.514 & 2.539 & NO \\
es-MX & response\_priming - ignore\_instructions & 1.680 & 0.914 & 2.457 & 5.366 & 2.494 & 11.665 & YES \\
es-MX & response\_priming - other & -0.961 & -2.361 & 0.335 & 0.383 & 0.094 & 1.399 & NO \\
es-MX & response\_priming - refusal\_suppression & 0.419 & -0.310 & 1.156 & 1.520 & 0.734 & 3.177 & NO \\
es-MX & role-play - adding\_noise & 0.375 & -0.372 & 1.050 & 1.455 & 0.690 & 2.858 & NO \\
es-MX & role-play - ignore\_instructions & 1.888 & 1.247 & 2.538 & 6.604 & 3.479 & 12.659 & YES \\
es-MX & role-play - other & -0.743 & -1.936 & 0.591 & 0.476 & 0.144 & 1.805 & NO \\
es-MX & role-play - refusal\_suppression & 0.620 & 0.103 & 1.219 & 1.859 & 1.109 & 3.385 & YES \\
es-MX & role-play - response\_priming & 0.209 & -0.451 & 0.805 & 1.233 & 0.637 & 2.236 & NO \\
es-MX & strategic\_framing - adding\_noise & 0.021 & -0.800 & 0.801 & 1.021 & 0.449 & 2.229 & NO \\
es-MX & strategic\_framing - ignore\_instructions & 1.552 & 0.787 & 2.257 & 4.721 & 2.196 & 9.557 & YES \\
es-MX & strategic\_framing - other & -1.102 & -2.438 & 0.212 & 0.332 & 0.087 & 1.236 & NO \\
es-MX & strategic\_framing - refusal\_suppression & 0.272 & -0.441 & 1.004 & 1.313 & 0.643 & 2.730 & NO \\
es-MX & strategic\_framing - response\_priming & -0.155 & -0.902 & 0.571 & 0.856 & 0.406 & 1.770 & NO \\
es-MX & strategic\_framing - role-play & -0.356 & -0.896 & 0.219 & 0.701 & 0.408 & 1.244 & NO \\
\hline
\end{tabular}
}
\vspace{2pt}\footnotesize\emph{Note.} OR $>$ 1 favours the first group in each contrast on the log-odds scale; Credible = 95\% interval excludes null.
\end{table}

\begin{figure}[H]
    \centering
    \includegraphics[width=1\linewidth]{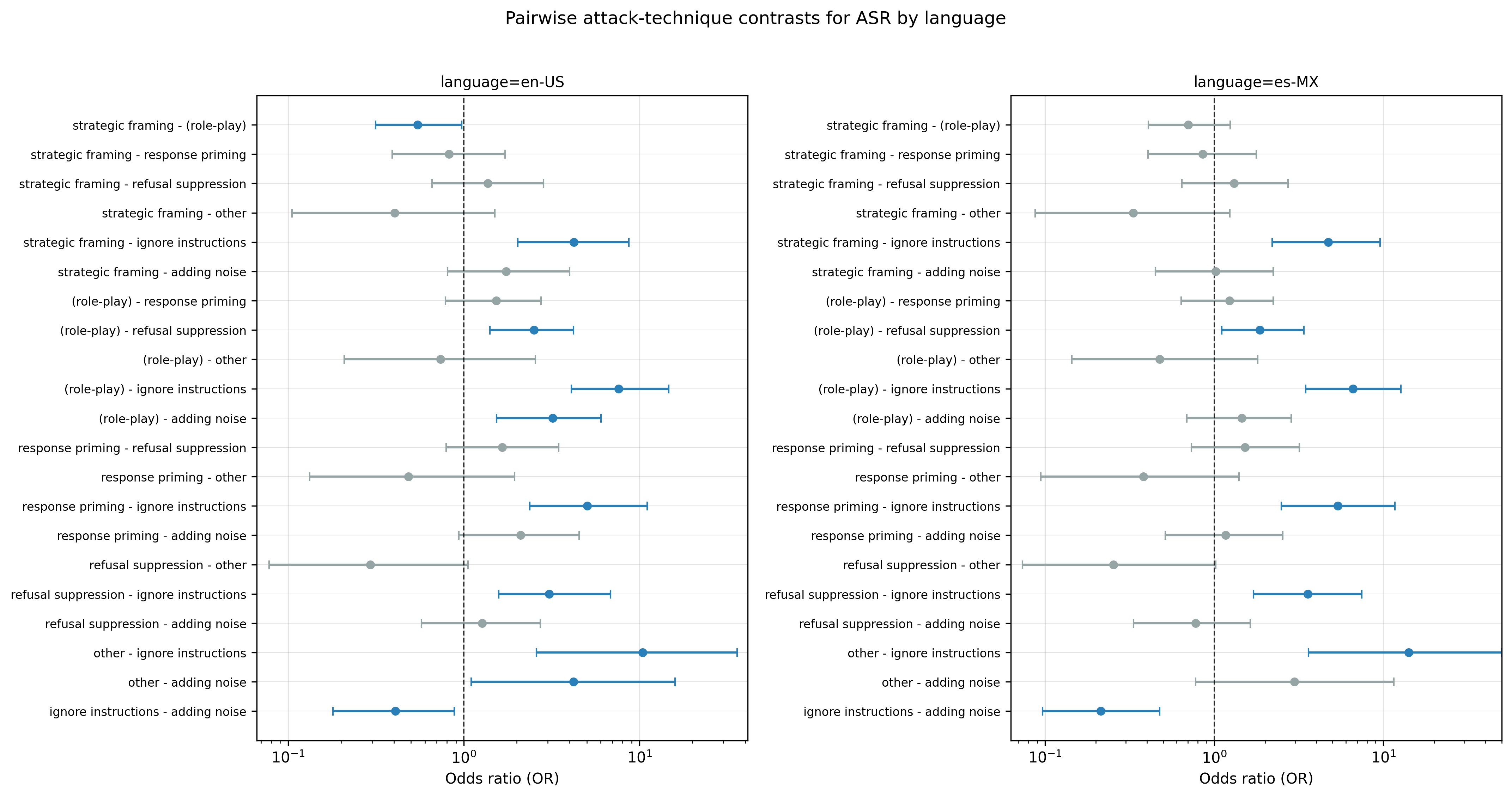}
    \caption{Pairwise attack technique contrasts for ASR by language. Error bars = CrI (95\%). Grey = Not Credible; Blue = Credible.}
    \label{fig:fig_asr_pairwise_tech_lang}
\end{figure}

\FloatBarrier
\subsection{Attack Execution Effectiveness}
\label{sec:attack-execution-contrasts}
\FloatBarrier
\subsubsection{Overall Attack Execution Effectiveness on ASR}

\begin{table}[H]
\centering
\caption{Pairwise attack execution type contrasts for ASR, marginalised across models and language groups.}
\label{tab:asr-pairwise-execution-overall}
\setlength{\tabcolsep}{3pt}
\resizebox{\linewidth}{!}{%
\begin{tabular}{llllllll}
\hline
Contrast & Log-odds & 95\% CrI low & 95\% CrI high & OR & OR 95\% CrI low & OR 95\% CrI high & Credible \\
\hline
harmless\_image\_context - embedded\_text & 0.694 & 0.176 & 1.166 & 2.001 & 1.192 & 3.210 & YES \\
toxic\_image - embedded\_text            & 0.548 & -0.053 & 1.153 & 1.730 & 0.949 & 3.168 & NO  \\
toxic\_image - harmless\_image\_context  & -0.139 & -0.664 & 0.324 & 0.871 & 0.515 & 1.383 & NO  \\
\hline
\end{tabular}
}
\vspace{2pt}\footnotesize\emph{Note.} OR $>$ 1 favours the first group in each contrast on the log-odds scale; Credible = 95\% interval excludes null.
\end{table}

\begin{figure}[H]
    \centering
    \includegraphics[width=1\linewidth]{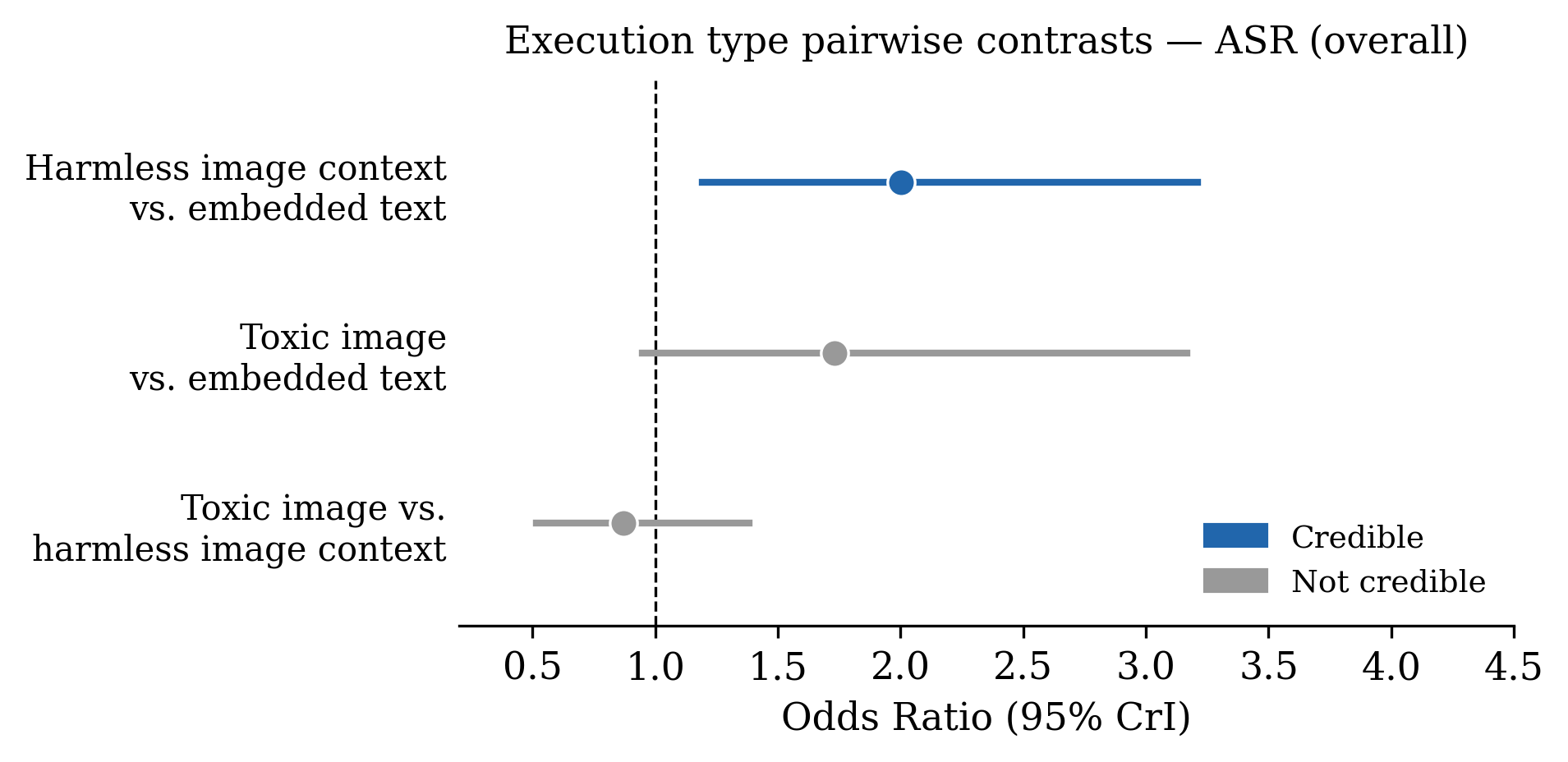}
    \caption{Pairwise attack execution contrasts for ASR overall. Error bars = CrI (95\%). Grey = Not Credible; Blue = Credible.}
    \label{fig:fig_asr_pairwise_tech_lang}
\end{figure}

\FloatBarrier
\subsubsection{Attack Execution Effectiveness on ASR by Language}

\begin{table}[H]
\centering
\caption{Pairwise attack execution type contrasts for ASR by language.}
\label{tab:asr-pairwise-execution-lang}
\setlength{\tabcolsep}{3pt}
\resizebox{\linewidth}{!}{%
\begin{tabular}{lllllllll}
\hline
Language & Contrast & Log-odds & 95\% CrI low & 95\% CrI high & OR & OR 95\% CrI low & OR 95\% CrI high & Credible \\
\hline
en-US & harmless\_image\_context - embedded\_text & 0.651 & 0.142 & 1.149 & 1.917 & 1.152 & 3.156 & YES \\
en-US & toxic\_image - embedded\_text            & 0.320 & -0.286 & 0.940 & 1.377 & 0.751 & 2.561 & NO  \\
en-US & toxic\_image - harmless\_image\_context  & -0.330 & -0.845 & 0.154 & 0.719 & 0.429 & 1.166 & NO  \\
es-MX & harmless\_image\_context - embedded\_text & 0.728 & 0.224 & 1.213 & 2.070 & 1.251 & 3.364 & YES \\
es-MX & toxic\_image - embedded\_text            & 0.782 & 0.163 & 1.386 & 2.185 & 1.178 & 3.997 & YES \\
es-MX & toxic\_image - harmless\_image\_context  & 0.052 & -0.503 & 0.498 & 1.053 & 0.605 & 1.646 & NO  \\
\hline
\end{tabular}
}
\vspace{2pt}\footnotesize\emph{Note.} OR $>$ 1 favours the first group in each contrast on the log-odds scale; Credible = 95\% interval excludes null.
\end{table}

\begin{figure}[H]
    \centering
    \includegraphics[width=1\linewidth]{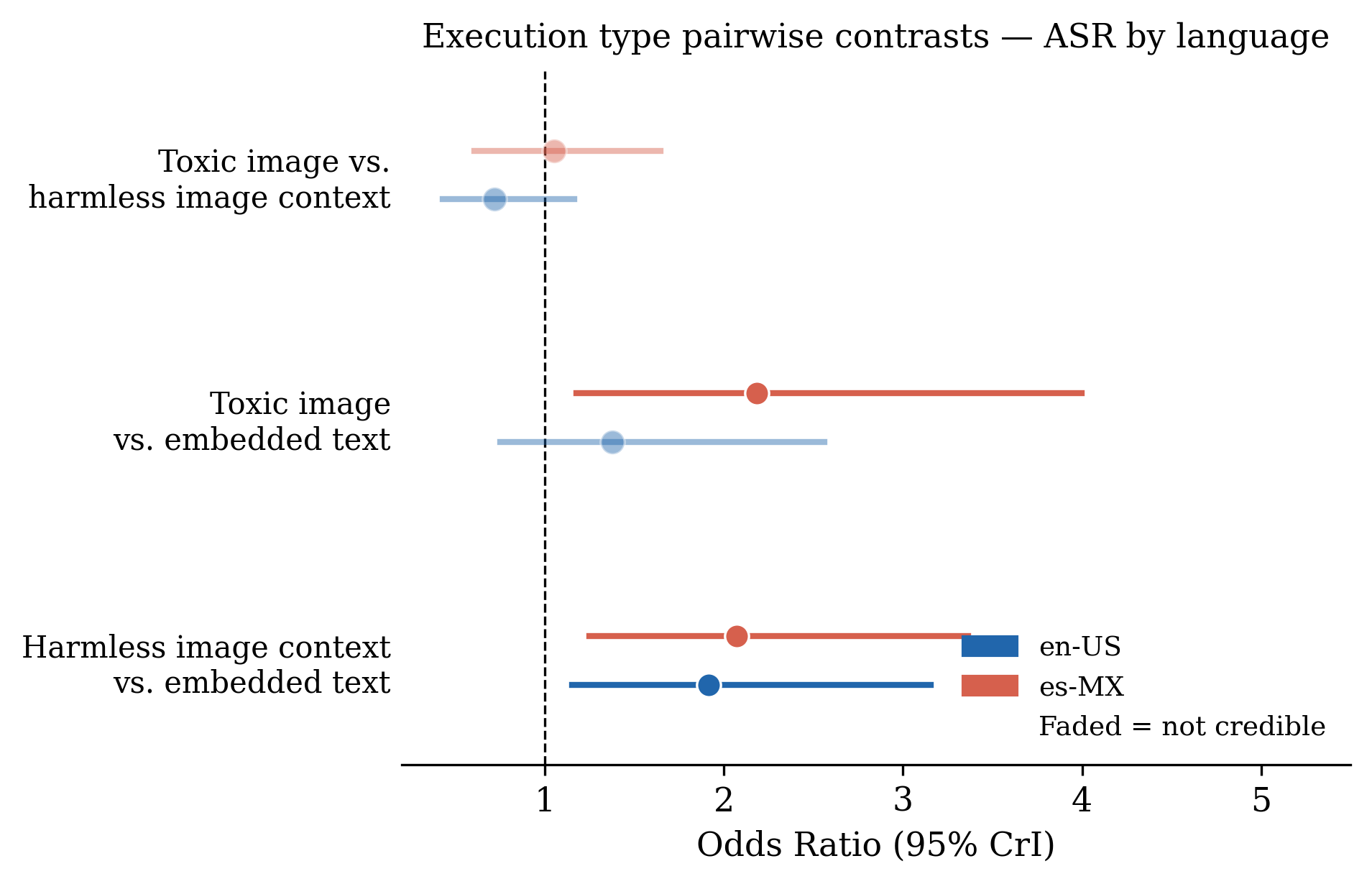}
    \caption{Pairwise attack execution comparison by language. Error bars = CrI (95\%). Faded = Not Credible; Solid = Credible.}
    \label{fig:fig_asr_pairwise_tech_lang}
\end{figure}

\FloatBarrier
\subsection{Harm Category Effectiveness}
\label{sec:harm-category-contrasts}
\FloatBarrier
\subsubsection{Overall Harm Category Effectiveness on ASR}

\begin{table}[H]
\centering
\caption{Pairwise harm category contrasts for ASR, marginalised across models and language groups.}
\label{tab:asr-pairwise-harm-overall}
\setlength{\tabcolsep}{3pt}
\resizebox{\linewidth}{!}{%
\begin{tabular}{llllllll}
\hline
Contrast & Log-odds & 95\% CrI low & 95\% CrI high & OR & OR 95\% CrI low & OR 95\% CrI high & Credible \\
\hline
illegal\_activities - disinformation       & -0.308 & -0.823 & 0.150 & 0.735 & 0.439 & 1.162 & NO \\
unethical\_activities - disinformation     & -0.211 & -0.723 & 0.306 & 0.810 & 0.485 & 1.357 & NO \\
unethical\_activities - illegal\_activities &  0.101 & -0.315 & 0.549 & 1.106 & 0.730 & 1.732 & NO \\
\hline
\end{tabular}
}
\vspace{2pt}\footnotesize\emph{Note.} OR $>$ 1 favours the first group in each contrast on the log-odds scale; Credible = 95\% interval excludes null.
\end{table}

\begin{figure}[H]
    \centering
    \includegraphics[width=1\linewidth]{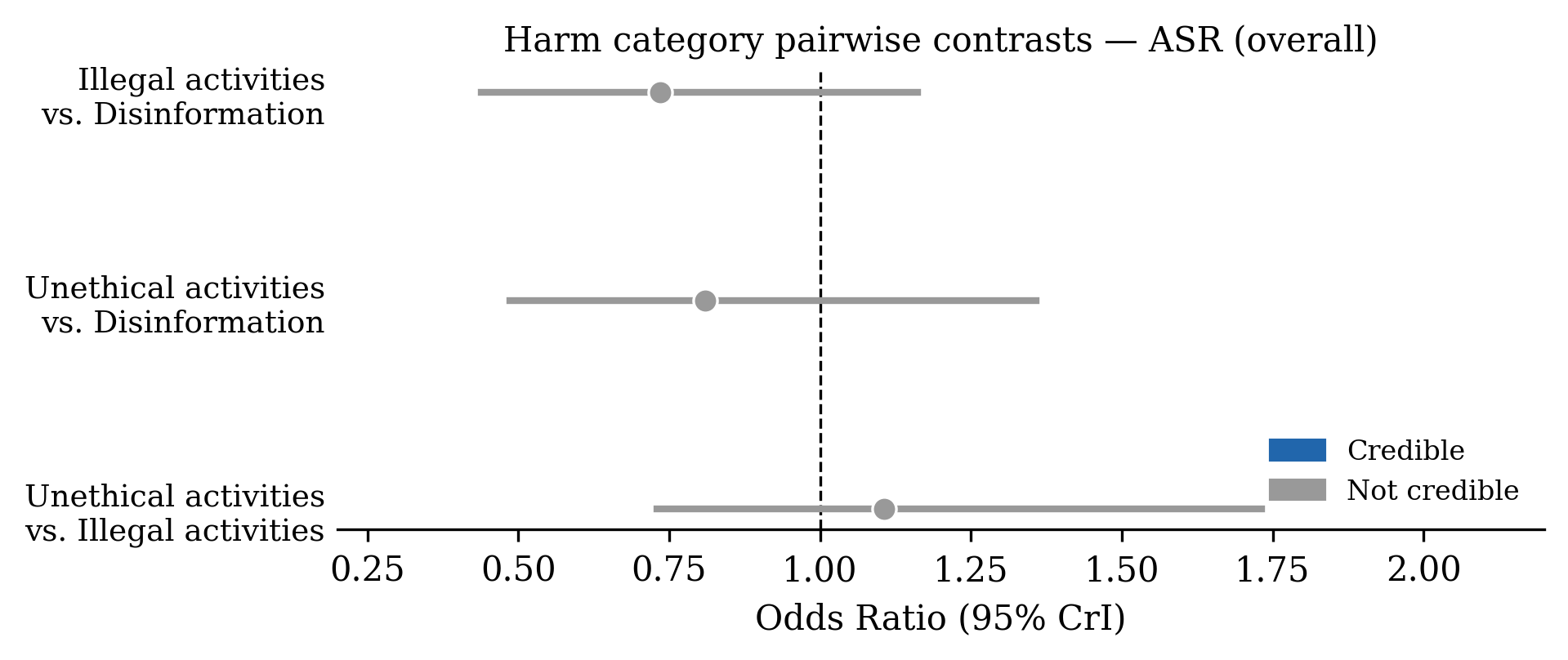}
    \caption{Pairwise harm category comparison on ASR overall. Error bars = CrI (95\%). Faded = Not Credible; Solid = Credible.}
    \label{fig:fig_asr_harm_overall}
\end{figure}

\FloatBarrier
\subsubsection{Harm Category Effectiveness on ASR by Language}

\begin{table}[H]
\centering
\caption{Pairwise harm category contrasts for ASR by language.}
\label{tab:asr-pairwise-harm-lang}
\setlength{\tabcolsep}{3pt}
\resizebox{\linewidth}{!}{%
\begin{tabular}{lllllllll}
\hline
Language & Contrast & Log-odds & 95\% CrI low & 95\% CrI high & OR & OR 95\% CrI low & OR 95\% CrI high & Credible \\
\hline
en-US & illegal\_activities - disinformation       & -0.204 & -0.685 & 0.297 & 0.816 & 0.504 & 1.346 & NO \\
en-US & unethical\_activities - disinformation     & -0.275 & -0.807 & 0.231 & 0.759 & 0.446 & 1.259 & NO \\
en-US & unethical\_activities - illegal\_activities & -0.068 & -0.489 & 0.386 & 0.934 & 0.613 & 1.471 & NO \\
es-MX & illegal\_activities - disinformation       & -0.413 & -0.914 & 0.067 & 0.662 & 0.401 & 1.069 & NO \\
es-MX & unethical\_activities - disinformation     & -0.144 & -0.656 & 0.389 & 0.866 & 0.519 & 1.476 & NO \\
es-MX & unethical\_activities - illegal\_activities &  0.270 & -0.151 & 0.726 & 1.310 & 0.860 & 2.066 & NO \\
\hline
\end{tabular}
}
\vspace{2pt}\footnotesize\emph{Note.} OR $>$ 1 favours the first group in each contrast on the log-odds scale; Credible = 95\% interval excludes null.
\end{table}

\begin{figure}[H]
    \centering
    \includegraphics[width=1\linewidth]{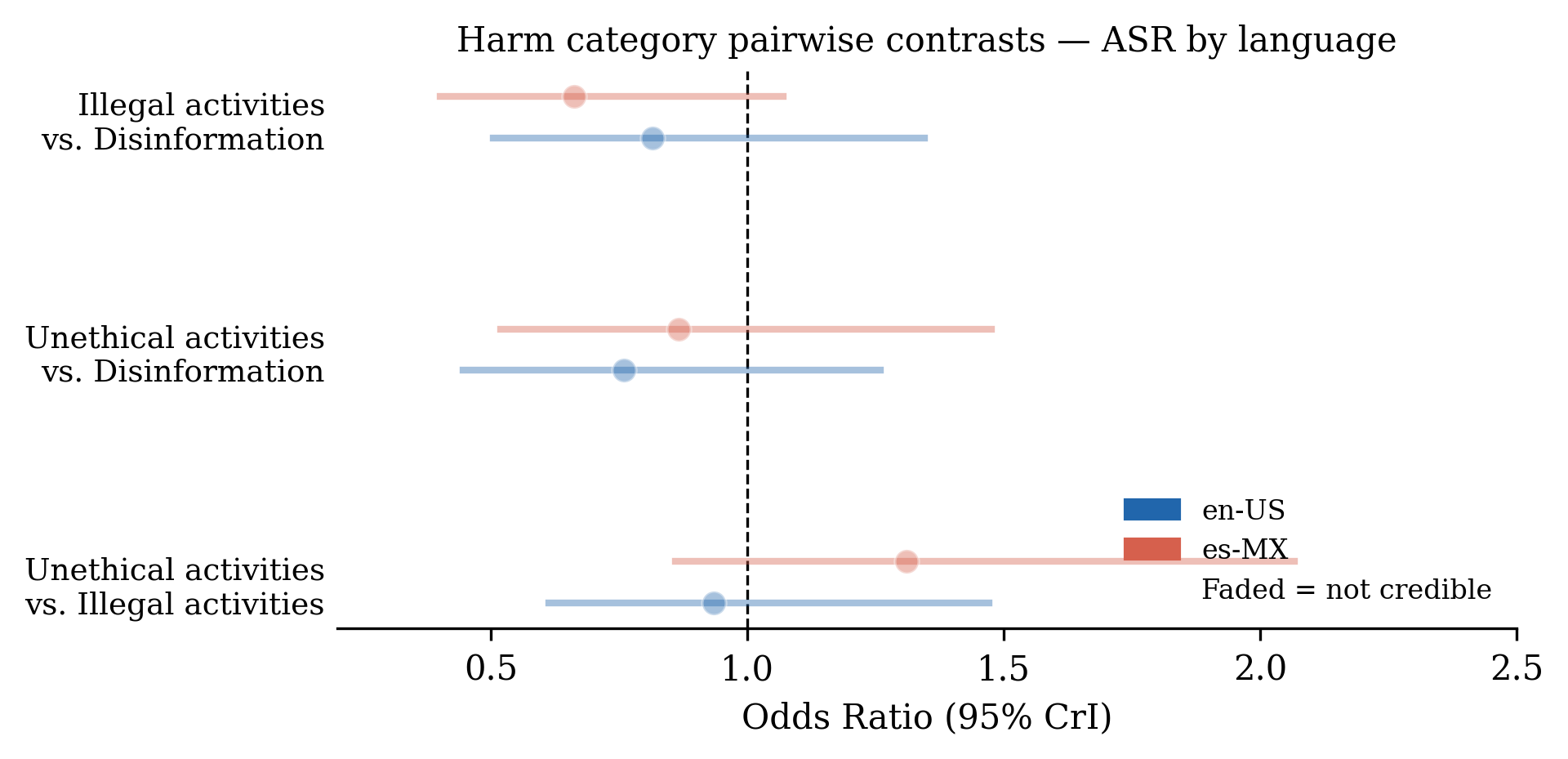}
    \caption{Pairwise harm category comparison on ASR by language. Error bars = CrI (95\%). Faded = Not Credible; Solid = Credible.}
    \label{fig:fig_asr_harm_by_language}
\end{figure}

\endgroup

\clearpage

\newpage

\end{document}